\documentclass{article}

\PassOptionsToPackage{numbers, compress, square}{natbib}
\usepackage[preprint]{neurips_2024}

\usepackage[utf8]{inputenc} % allow utf-8 input
\usepackage[T1]{fontenc}    % use 8-bit T1 fonts
\usepackage{hyperref}       % hyperlinks
\usepackage{url}            % simple URL typesetting
\usepackage{booktabs}       % professional-quality tables
\usepackage{amsfonts}       % blackboard math symbols
\usepackage{nicefrac}       % compact symbols for 1/2, etc.
\usepackage{microtype}      % microtypography
\usepackage{xcolor}         % colors
\usepackage{graphicx}
\usepackage{booktabs}
\usepackage{tcolorbox}
\usepackage{amsmath}
\usepackage{wrapfig}
\usepackage{subcaption}
\usepackage{multirow}
\usepackage{siunitx}
\usepackage{titletoc}
\usepackage[table]{xcolor}
\definecolor{cahgreen}{RGB}{198, 239, 206}
\definecolor{cmrblue}{RGB}{189, 215, 238}
\usepackage{CJKutf8}
\usepackage{cleveref}

\title{Counting Circuits: Mechanistic Interpretability of Visual Reasoning in Large Vision-Language Models}

\author{%
  \textbf{Liwei Che}$^{*,1}$ \quad
  \textbf{Zhiyu Xue}$^{*,2}$ \quad
  \textbf{Yihao Quan}$^{*,1}$ \quad
  \textbf{Benlin Liu}$^{3,4}$ \quad
  \textbf{Zeru Shi}$^{1}$ \quad
  \textbf{Michelle Hurst}$^{1}$ \\
  \textbf{Jacob Feldman}$^{1}$ \quad
  \textbf{Ruixiang Tang}$^{1}$ \quad
  \textbf{Ranjay Krishna}$^{3,4}$ \quad
  \textbf{Vladimir Pavlovic}$^{1}$ \\[4pt]
  $^{1}$Rutgers University \quad
  $^{2}$UC Santa Barbara \quad
  $^{3}$University of Washington \quad
  $^{4}$Allen Institute for AI \\[2pt]
  \texttt{\{lc1279, yq207, zs618, rt836, vladimir\}@cs.rutgers.edu} \\
  \texttt{michelle.hurst@rutgers.edu} \quad
  \texttt{jacob@ruccs.rutgers.edu} \\
  \texttt{zhiyuxue@ucsb.edu} \quad
  \texttt{liubl@cs.washington.edu} \quad
  \texttt{ranjay@cs.washington.edu} \\[2pt]
  \small $^{*}$Equal contribution
}

\begin{document}

\maketitle

\begin{abstract}

Counting serves as a simple but powerful test of a Large Vision-Language Model's (LVLM) reasoning; it forces the model to identify each individual object and then add them all up.
In this study, we investigate how LVLMs implement counting using controlled synthetic and real-world benchmarks, combined with mechanistic analyses. 
Our results show that LVLMs display a human-like counting behavior, with precise performance on small numerosities and noisy estimation for larger quantities.
We introduce two novel interpretability methods, Visual Activation Patching and HeadLens, and use them to uncover a structured “counting circuit” that is largely shared across a variety of visual reasoning tasks. Building on these insights, we propose a lightweight intervention strategy that exploits simple and abundantly available synthetic images to fine-tune arbitrary pretrained LVLMs exclusively on counting.  Despite the narrow scope of this fine-tuning, the intervention not only enhances counting accuracy on in-distribution synthetic data, but also yields an average improvement of +8.36\% on out-of-distribution counting benchmarks and an average gain of +1.54\% on complex, general visual reasoning tasks for Qwen2.5-VL. These findings highlight the central, influential role of counting in visual reasoning and suggest a potential pathway for improving overall visual reasoning capabilities through targeted enhancement of counting mechanisms.

%Counting requires explicit instance individuation and aggregation, making it a minimal yet revealing probe of visual reasoning for Large Vision Language Models (LVLMs). In this work, we study how LVLMs perform counting through controlled synthetic and real-world benchmarks with mechanistic analysis. We reveal that LVLMs exhibit similar visual counting characteristics to humans. By introducing Visual Activation Patching and HeadLens, two novel interpretability techniques, we identify a structured counting circuit that is largely shared by diverse visual reasoning tasks.
%Guided by these insights, we propose a lightweight intervention method that takes advantage of the simplest and most abundant synthetic images to fine-tune any pretrained LVLM on the counting task only. Surprisingly, this method not only improves the counting performance on in-distribution synthetic data, but also brings about $+8.36\%$ on out-of-distribution counting and $+1.54\%$, on average, on complex general visual reasoning tasks. This suggests the central and impactful role that counting plays in visual reasoning tasks, as well as a possible mechanism to increase general visual reasoning performance.

\end{abstract}

\section{Introduction}

% budget 1 page

Counting is one of the most fundamental yet revealing capacities of visual intelligence. Cognitive science reveals that human numerosity perception is shaped by severe information-processing constraints, resulting in near-perfect accuracy for small sets (subitizing) and noisy estimation for larger quantities \cite{cheyette2020unified}. Recent work further demonstrates that these discontinuities arise not from separate cognitive modules, but from resource-rational trade-offs~\cite{lieder2020resource} between precision, exposure time, and environmental statistics, suggesting that counting reflects a core organizing principle of visual reasoning rather than a task-specific heuristic~\cite{cheyette2024limited}.

Recent LVLMs~\cite{liu2023visual,bai2025qwen2} have demonstrated remarkable generalization across diverse visual reasoning tasks~\cite{yue2024mmmu,lu2023mathvista,grok15v2024xai,wang2024measuring}. However, our benchmarks (\cref{tab:initial_counting}) reveal a surprising phenomenon: these models struggle to count even fewer than ten simple black dots, a trivial task for human vision. This discrepancy raises fundamental questions regarding the nature of counting ability in LVLMs and its relation to general visual reasoning. Unlike standard recognition, counting cannot be bypassed through semantic memorization or dataset priors. It strictly requires models to individuate discrete entities, maintain intermediate representations, and aggregate information under architectural constraints. 
Therefore, in this work, we utilize the counting task as a minimalist probe to investigate the internal visual reasoning process of LVLMs and explore how these mechanism discoveries can support more complex, high-level visual reasoning tasks.

\begin{figure}[!t]
    \centering
    \includegraphics[width=1\linewidth]{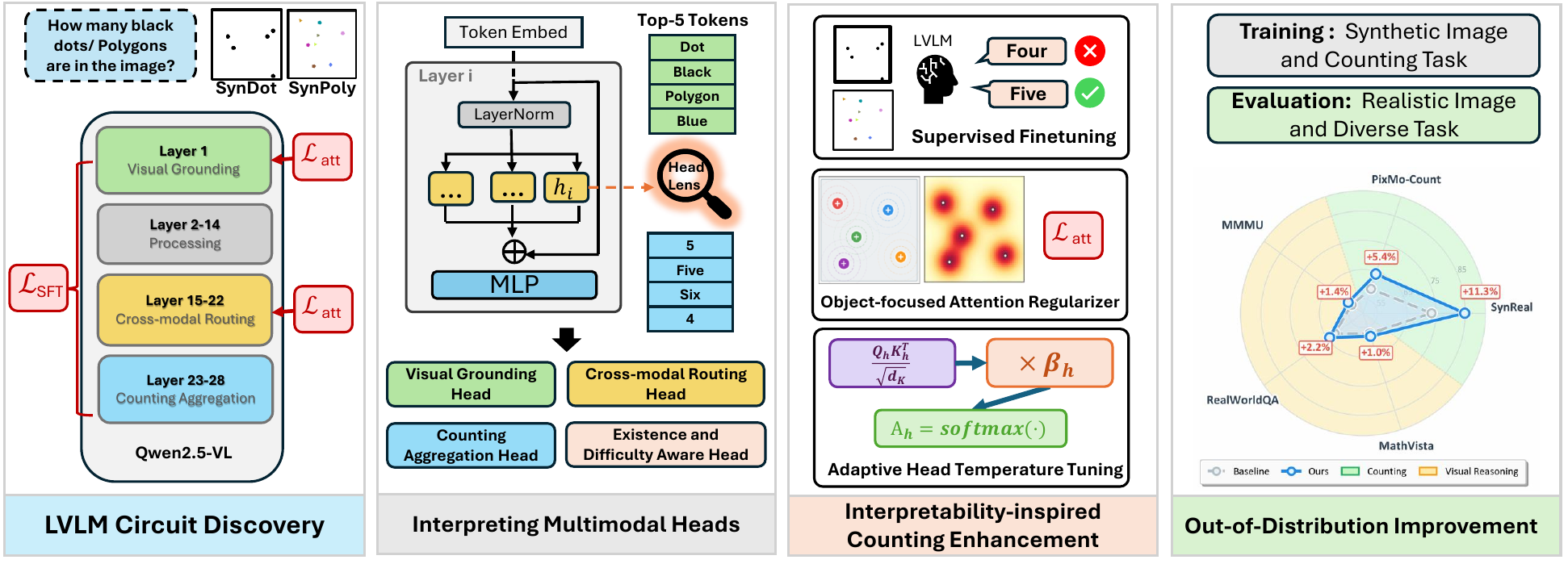}
    \caption{Overview of Main Contributions}
    \label{fig:main}
\end{figure}

Existing works often treat counting as an isolated task, attempting to improve the counting performance of LVLMs with methods like image preprocessing~\cite{qharabagh2024lvlm}, fine-tuning~\cite{paiss2023teaching}, and attention intervention~\cite{sengupta2025can}. However, as specialized computer vision methods~\cite{Dukic_2023_ICCV,Huang_2024_CVPR} undoubtedly yield better performance, we argue that understanding and evaluating the visual reasoning process of LVLMs via counting should be the primary focus. Closer work such as CountScope~\cite{hasani2025understanding} and [De|Re]~\cite{alghisi2025re} heavily rely on probing methods to understand the counting pattern of LVLMs, which is unreliable due to the simplicity of the representation of their synthetic data. More importantly, they fail to understand the circuit-level mechanism of counting.
Therefore, in this work, we investigate how LVLMs execute counting tasks from the dual perspectives of human cognitive alignment and mechanistic interpretability. First, we leverage the controllability of synthetic images to extend activation patching~\cite{heimersheim2024use} to multimodal inputs. By tracing the flow of counting-related information across model layers, we reveal the cross-modal information routing occurring in the middle layers and the emergence of counting answers in the later layers. To further isolate and verify the roles of different model components, we propose HeadLens, a novel tool that interprets the output of attention heads into semantic tokens.
Through this, we identify four critical categories of attention heads: (1) \textit{Visual Grounding Heads} (Early Layers): Extract foundational visual features (color, shape) directly from image tokens. (2) \textit{Cross-Modal Routing Heads} (Middle Layers): Translate visual information into abstract numerical concepts. (3) \textit{Counting Aggregation Heads} (Late Layers): Attend heavily to text prompts; their top-10 HeadLens decoded tokens highly correlate with the final prediction. (4) \textit{Awareness Heads} (Late Layers): Two specialized heads encoding object existence and difficulty estimations.

Building on the interpretability findings, we design targeted intervention methods tailored to specific layers and heads via parameter tuning, attention regularization, and head temperature tuning. Utilizing 8,000 highly simplified synthetic images (e.g., black dots or colored polygons on a white canvas) that can be generated in minutes on a standard computer, our approach not only improves performance on synthetic data but also achieves up to an $8.36\%$ average accuracy increase on out-of-distribution (OOD) real-world object counting. Furthermore, it yields a $1.54\%$ average accuracy improvement across broader OOD visual tasks (e.g., general VQA, visual math). This demonstrates that counting serves as a vital foundational pillar of general visual intelligence and validates the existence of generalized visual reasoning mechanisms operating as internal circuits within LVLMs.
Our main contributions are summarized as follows:

\noindent\textbf{Cognitive Alignment:} Our investigation reveals that LVLMs exhibit a striking discontinuity in visual counting that mirrors human behavior, characterized by precise subitizing for small sets and noisy estimation for larger quantities. We ground the model's behavioral failures in the topological entanglement of hidden state manifolds.

\noindent\textbf{Methodological Innovation:} We propose {Visual Activation Patching} (VAP) and {HeadLens}, two novel mechanistic interpretability techniques designed to isolate head-specific functions and perform circuit discovery for LVLMs.

\noindent\textbf{Circuit Discovery:} We identify four distinct functional categories of attention heads essential for visual counting: \textit{visual grounding}, \textit{cross-modal routing}, \textit{counting aggregation}, and \textit{difficulty/existence detection}. We further demonstrate that a great proportion of these specialized heads also serve as foundational components for general visual reasoning.

\noindent\textbf{Performance Enhancement:} We develop an interpretability-inspired intervention strategy. Using only the simplest synthetic images, we significantly enhance the model's OOD counting robustness and its performance on complex, general visual reasoning benchmarks.

\section{Related Work}
\label{sec:related_work}

\subsection{LVLMs for counting task.} 
The strong general visual capabilities of LVLMs have sparked considerable attention to their counting performance. 
Early exploration~\cite{paiss2023teaching} utilizes additional counting data to fine-tune the CLIP for better counting performance.
LVLM-count~\cite{qharabagh2024lvlm} utilizes SAM~\cite{kirillov2023segment} and a pixel search algorithm to preprocess the image into pieces for better counting performance. 
Guo et al.\cite{guo2025your} explore the relationship between object types and quantities, revealing the compositional counting failure modes of LVLMs. More recent works explore the counting task via internal states analysis.
CountScope~\cite{hasani2025understanding} verified that LVLMs will accumulate counting information along the layers and tokens in a causal order due to the auto-regressive pattern, yet failed to provide circuit-level functionality understanding on how the internal mechanism actually works. [De|Re]~\cite{alghisi2025re} identified that the incorrect mapping of the last layer caused the counting error, while ignoring that the simplistic representational structure of the synthetic data renders probing results unreliable. An over $99\%$ counting accuracy by linear probing can be trivially achieved even before the LVLM processes the input. More importantly, they failed to reveal the functions of architecture components such as attention heads and layers which are the fundamental building blocks of the transformer's computation. Sengupta et al.~\cite{sengupta2025can} propose attention-based interventions that redistribute attention weights to improve counting, but their analysis is limited to behavioral outcomes without dissecting which heads carry counting-specific signals or how they interact across layers. In contrast, our work goes beyond layer-level probing and behavioral intervention: we perform head-level circuit discovery via Visual Activation Patching and HeadLens, identify four functionally distinct head categories, and demonstrate that the discovered counting circuit generalizes to broader visual reasoning tasks.

\subsection{Mechanistic Interpretability on LLM/LVLM.}
Prior work on interpreting LLM/VLM internals on the layer level has explored probing the direct relationship between intermediate hidden states and final model outputs. Logit Lens~\cite{nostalgebraist2020logitlens} proposes to project hidden representations at various layers directly into the vocabulary space using the model’s output head. It can reveal how information related to the next token emerges throughout layers. Tuned Lens~\cite{belrose2023eliciting} extends logit-based probing by learning a linear translator from hidden states to logits. It improves the fidelity of the probe but still operates at the level of entire layer activations. 
Existing head-level interpretability methods like AttentionLens~\cite{sakarvadia2023attention} require training independent translators for each attention head. It is computationally expensive and risks masking the specialized role of a head by forcing it to mimic the final output. HeadLens overcomes this by leveraging the block-matrix additivity of the attention projection. It ensures the semantic space of each attention head, and reveals the authentic contribution of each head to the residual stream.

Beyond representation probing, recent advances in mechanistic interpretability employ causal interventions to isolate the functional roles of specific components. For instance, Meng et al.~\cite{meng2022locating} introduce causal tracing to identify that mid-layer feed-forward networks (MLPs) are decisive in recalling factual associations, which enables direct model editing techniques such as Rank-One Model Editing (ROME). 
Similarly, Wang et al.~\cite{wang2022interpretability} apply causal interventions to reverse-engineer a comprehensive attention circuit for the Indirect Object Identification (IOI) task in GPT-2, successfully categorizing specific attention heads into specialized functional classes (e.g., name mover heads, induction heads). 
These works highlight the necessity of isolating individual component contributions rather than treating layer activations as a monolith, further motivating our fine-grained, head-level analysis.

In the realm of Large Vision-Language Models (LVLMs), mechanistic interpretability is actively expanding to understand cross-modal interactions and visual token processing. Neo et al.~\cite{neo2025towards} investigate how visual information evolves within the language model backbone of LLaVA, demonstrating that visual token representations gradually align with interpretable textual concepts in the vocabulary space across deeper layers. Furthermore, they reveal that the model extracts this localized object information at the final token position for prediction, mirroring factual recall in text-only models. 
Additionally, fine-grained analysis of attention mechanisms has proven crucial for diagnosing model failures~\cite{Che_2025_ICCV}.  Collectively, these studies underscore that understanding token-level routing and head-level attention mechanisms is essential for interpreting and improving LVLM behaviors.
\section{Problem Formulation}

% To validate the intrinsic counting capabilities of LVLMs, we employ synthetic datasets designed to isolate numerical reasoning from confounding variables such as background clutter, perspective distortions, and complex textures. 
% As in~\cref{fig:syndata}, we use PIL~\cite{clark2015pillow} generated \textbf{SynDot} and \textbf{SynPoly}, comprising randomly distributed black dots or multi-colored polygons on a white canvas; and \textbf{SynReal}, which utilizes FLUX.1-dev~\cite{labs2025flux1kontextflowmatching} to generate photorealistic objects across six common object categories. We limited the number range of objects to $[1,10]$ for all three benchmarks and describe more details in~\cref{sec:dataset}.

% For a comprehensive assessment, we evaluate the model's counting performance across four distinct axes (formally defined in~\cref{sec:eval_metric}), including \textbf{Accuracy (ACC)} for exact-match precision; \textbf{Mean Absolute Error (MAE)} and \textbf{Root Mean Square Error (RMSE)} to capture the magnitude and stability of numerical deviations; \textbf{Off-by-one Accuracy (OBO)} to assess "near-miss" reliability, a proxy for human-like subitizing.

To validate the intrinsic counting capabilities of LVLMs, we employ synthetic datasets that isolate numerical reasoning from confounding variables such as background clutter and complex textures. 
As in~\cref{fig:main}, we use PIL~\cite{clark2015pillow} generated \textbf{SynDot} and \textbf{SynPoly}, comprising randomly distributed black dots or multi-colored polygons on a white canvas; and \textbf{SynReal}, which uses FLUX.1-dev~\cite{labs2025flux1kontextflowmatching} to generate photorealistic objects across six categories. Object counts range in $[1,10]$; details are in~\cref{sec:dataset}.
We evaluate counting performance on four axes (formally defined in~\cref{sec:eval_metric}): \textbf{Accuracy (ACC)} for exact-match precision; \textbf{MAE} and \textbf{RMSE} for deviation magnitude and stability; and \textbf{Off-by-one Accuracy (OBO)} for near-miss reliability.
% Collectively, this suite of metrics transcends simple error measurement by decoupling precision from robustness and confidence. 

\begin{table}[ht]
\centering
\caption{Performance comparison on synthetic benchmarks.}
\label{tab:initial_counting}
\small 
\setlength{\tabcolsep}{5pt} 
\begin{tabular}{llcccc}
\toprule
\textbf{Model} & \textbf{Dataset} & \textbf{Acc} $\uparrow$ & \textbf{MAE} $\downarrow$ & \textbf{RMSE} $\downarrow$ & \textbf{OBO} $\uparrow$ \\ \midrule
\multirow{3}{*}{Qwen2.5-VL-7B} 
& SynDot  & 76.12 & 0.25 & 0.52 & 96.21 \\
& SynPoly & 53.74 & 0.94 & 1.75 & 82.27 \\
& SynReal & 73.48 & 0.79 & 2.13 & 91.01 \\ \midrule

\multirow{3}{*}{Qwen3-VL-8B}   & SynDot  & 73.22 & 0.32 & 0.66 & 95.21 \\
                               & SynPoly & {70.34} & {0.52} & 1.49 & {94.02}  \\
                               & SynReal & 88.79 & 0.15 & 0.47 & {98.88}  \\ \midrule

\multirow{3}{*}{LLaVA-1.5-7B}  
& SynDot  & 24.07 & 4.46 & 6.58 & 39.22 \\
& SynPoly & 33.30 & 1.65 & 2.34 & 56.71 \\
& SynReal & 54.27 & 3.59 & 5.05 & 66.34 \\ 
\bottomrule
\end{tabular}
\end{table}

We evaluate the counting performance of three popular LVLMs, with the prompt "\textit{What is the number of the \{object\} in the image? Answer with number only.}". As shown in \cref{tab:initial_counting}, we surprisingly find that these models fail significantly across all three benchmarks compared to humans, who can achieve flawless accuracy. A more counterintuitive observation is that LVLMs perform better on the more realistic SynReal, while struggling with the visually simpler SynDot and SynPoly. This discrepancy clearly exposes an inherent bias rooted in their training data distribution. This naturally raises a fundamental question: \textbf{Does the counting capability of LVLMs signify an emergent reasoning mechanism, or is it merely a statistical hallucination predicated on correlational visual cues and language priors? }

% \RX{Using GPT-5 to parse model responses introduces a dependency on a black-box system. Reviewers may question the reproducibility. Consider (1) reporting the parsing agreement rate or providing a few failure examples, and (2) adding a simple regex-based parser as a baseline to show GPT-5 parsing is reliable.''}

% \RX{Table 1: The MAE and RMSE values seem disproportionately large for a 1--10 range (e.g., MAE=446.23 for LLaVA SynDot). This suggests the model sometimes outputs very large numbers or the penalty value $p_i=-1$ is inflating these metrics. Clarify how the penalty value $-1$ interacts with MAE/RMSE computation---are non-answers excluded or counted as error? If counted, this conflates formatting failures with counting failures. Consider reporting MAE/RMSE only on the subset where AR=100\% (i.e., valid answers only).}

\section{Do LVLMs know how to count? Or do they memorize?}
\label{sec:if_count}

\begin{figure}
    \centering
    \includegraphics[width=1\linewidth]{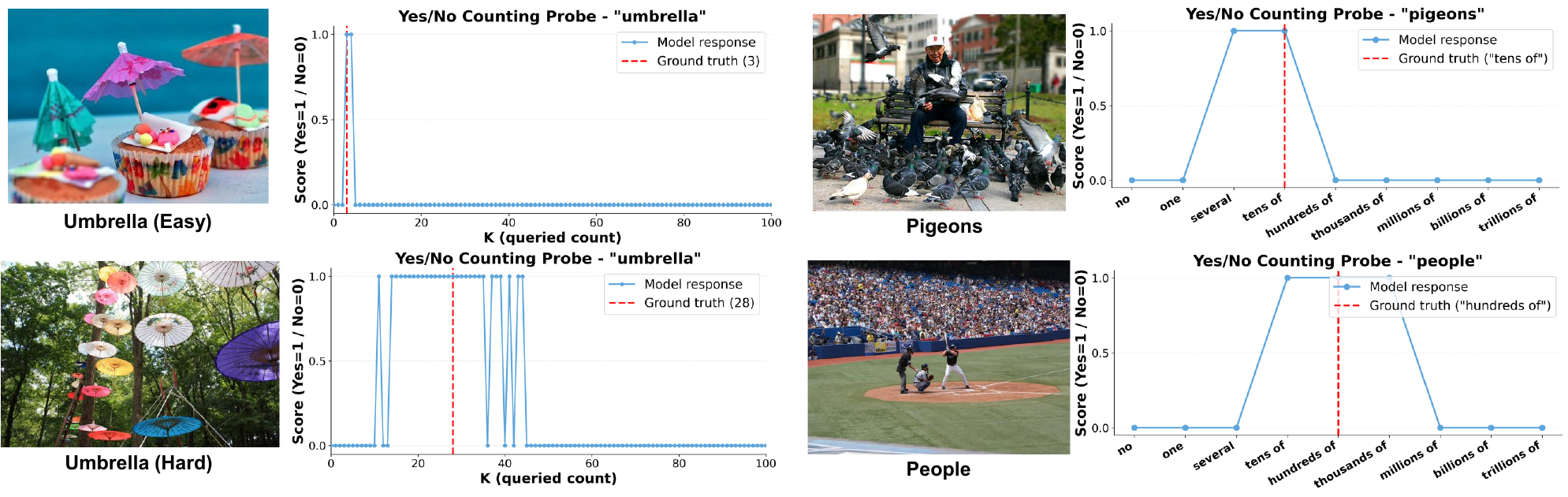}
    \caption{Counting Uncertainty Curve by Yes/No Answer of Qwen2.5 VL 7B. }
    \label{fig:if_count}
\end{figure}

In this section, we attempt to address the problem proposed in the initial evaluation.
We observe that models often hesitate to provide precise numerical counts when faced with occlusion, blurriness, or high object density. 
To circumvent this, we reformulate the counting task into a binary 'Yes/No' verification framework. 
Specifically, for a given image, we employ the prompt: ``\textit{There are <K> <Object Name> in the image, Yes or No?}". 
For images with countable instances, we iteratively vary $K$ from $0$ to $100$. For complex scenes (e.g., swarms of pigeons or dense crowds), we substitute $K$ with quantitative descriptors such as 'tens of' or 'hundreds of'. 
We assign scores of $1$ for 'Yes', $0$ for 'No'.
% \RX{The Yes/No probing is clever, but the score of 0.5 for ambiguous responses is ad hoc. How frequent are ambiguous responses? If rare, they do not matter; if common, the 0.5 assignment materially shapes the curves. Report the ambiguity rate and consider a sensitivity analysis with different ambiguous scores (e.g., 0 and 1) to show the curves are robust.}

% 

As shown in~\cref{fig:if_count}, counting uncertainty strongly correlates with the stability of the model's ``Yes Band'' (a contiguous interval of positive responses). Simple scenarios (e.g., few, unoccluded umbrellas) yield a concentrated, stable Yes Band, whereas challenging cases blur decision boundaries, causing the band to expand and oscillate. Nonetheless, LVLMs retain robust coarse-grained estimation, reliably distinguishing orders of magnitude (e.g., ``tens'' vs. ``millions'' of pigeons) despite lacking precision. 
These properties validate that: \textit{LVLMs possess a visual subitizing and estimation capability similar to human behavior~\cite{cheyette2020unified}. Instead of relying on discrete tokens, the model encodes quantity within a continuous latent space, enabling seamless cross-modal alignment between visual stimuli and linguistic quantifiers.}

% \RX{The analogy to human subitizing is compelling but may be over-claimed. Human subitizing is defined by reaction-time experiments (fast, parallel processing for $\leq$4 items), not just accuracy. LVLMs do not have ``reaction time'' in the same sense. Consider softening the claim to ``behavioral similarity'' rather than ``analogous capability,'' or provide additional evidence such as showing that the number of forward-pass layers needed to resolve counts 1--4 is fewer than for 5--10 (which would be a more mechanistic parallel to the speed argument).}

%  \begin{tcolorbox}[width=\textwidth,colback={grey}]  
%    LVLMs possess a visual subitizing and estimation capability analogous to human cognition~\cite{cheyette2020unified}. Instead of relying on discrete tokens, the model encodes quantity within a continuous latent space, enabling seamless cross-modal alignment between visual stimuli and linguistic quantifiers. 
% \end{tcolorbox} 

\begin{figure}
    \centering
    \includegraphics[width=0.75\linewidth]{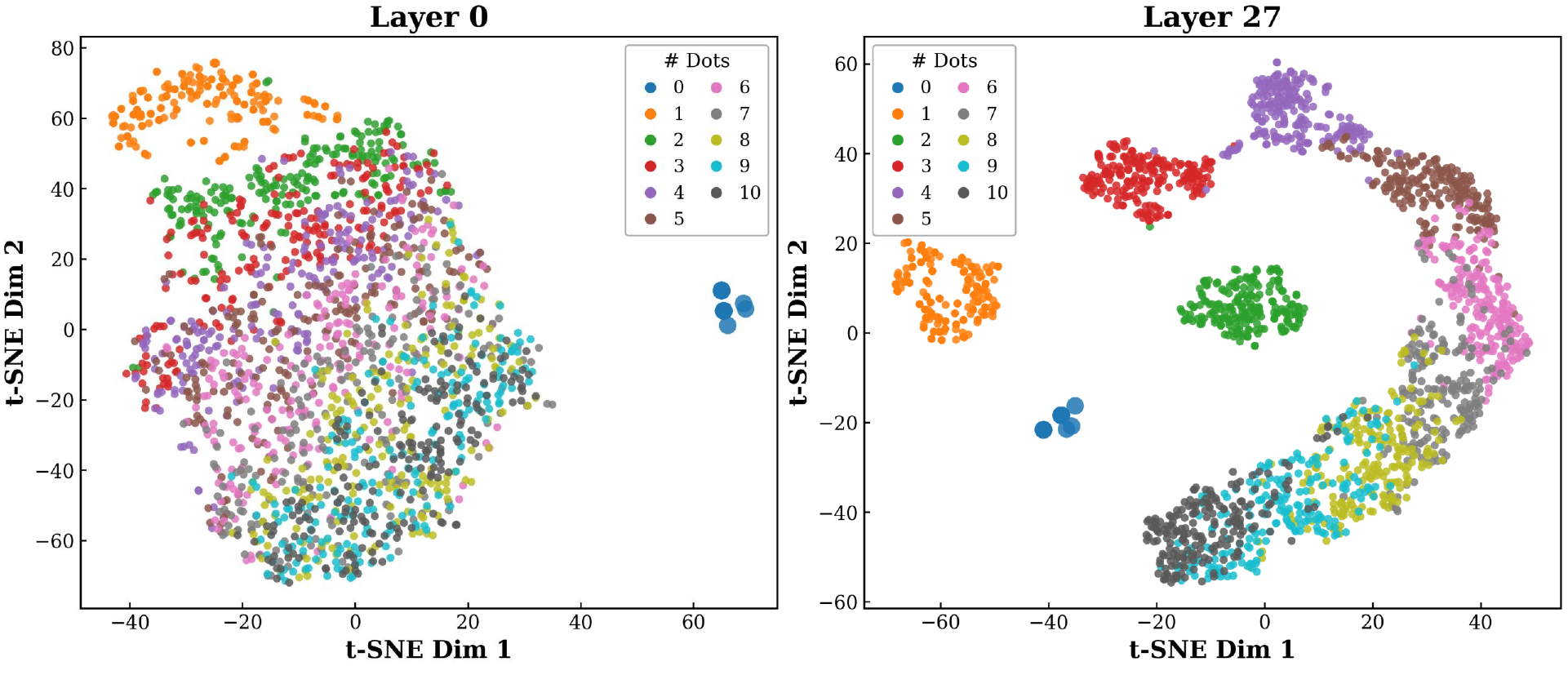}
    \caption{T-SNE visualization of model hidden states based on black dots data.}
    \label{fig:tsne}
\end{figure}

\noindent\textbf{Cognitive Alignment from the Representation Aspect.}
To further verify the counting ability alignment between LVLMs and human cognition, we project the hidden states of the SynDot dataset into a 2D space using t-SNE (\cref{fig:tsne}). As features propagate from the first to the final layer, \textit{class 0} (background) and \textit{class 1 - 4} (black dots number) form distinct, isolated clusters separated from other numbers. 
This topological isolation provides a mechanistic explanation for the model's subitizing capability, the accurate recognition of small quantities. However, as the object count increases, the corresponding clusters become progressively entangled and heavily overlapped. This spatial compression in the latent space directly accounts for the performance degradation on larger numbers, reflecting a natural cognitive shift from precise subitizing to approximate magnitude estimation.

% \RX{The t-SNE visualization (Figure 3) is qualitative and depends heavily on perplexity and random seed. Consider supplementing with a quantitative clustering metric (e.g., silhouette score, inter-/intra-cluster distance ratio) across layers to make the ``progressive entanglement'' claim rigorous. Also, t-SNE is shown for SynDot only---does the same pattern hold for SynPoly and SynReal? If yes, showing this in the appendix would strengthen the generality claim. If not, that is also interesting to discuss.}

\section{Understand Layerwise Behavior of Counting}

In this section, we investigate the underlying mechanisms that execute visual counting tasks inside LVLMs.  
We choose Qwen2.5-VL-7B as the default model for our interpretability studies due to its balance between architectural simplicity and good counting performance.

\subsection{Information Flow and Cross-Modal Routing}
\label{sec:information_flow}

To trace how counting information propagates from visual input to numerical output, we adapt activation patching~\cite{heimersheim2024use,zhang2023towards}, a technique traditionally applied to text-only LLMs~\cite{yeo2025towards,dumas2025separating}, for LVLM analysis. By fixing the random seed in our synthetic datasets, we construct tightly controlled image pairs. Each pair consists of a clean image (e.g., three dots) and a corrupted counterpart with an altered object count (e.g., five dots) that strictly preserves the original spatial layout. This minimal perturbation extends activation patching to the visual modality, allowing us to precisely isolate how the model's internal states respond to quantitative changes.

% \noindent\textbf{Visual Activation Patching.} 
%  For an LVLM with $L$ layers, the hidden state $s^l_i$ carries the information of $x_i$ processed by the previous $l$ layers and will be fed to the $l+1$ th layer. 
% We first capture the hidden states generated during an inference pass on the corrupted image. 
% Subsequently, during the inference of the clean image, we perform an intervention at a target layer $l$ by replacing the hidden states $s^l_i$ of a specific token or token range with their corresponding representations from the corrupted run.
% By observing whether the model's final count changed from the clean image label to the corrupted label, we calculate the \textit{overwrite rate}, which measures the causal efficacy of the patched tokens.
% We categorize input tokens into six functional groups: System Prompt, Image Tokens, User Instruction, Generated Tokens, Last Image Token(``|img\_end|"), and Last Prompt Tokens (the ``Assistant:" role tag). We set the prompt as "\textit{What is the number of the black dots in the image? Answer with the number only.}" such that the first generated token will be the target counting answer.

\noindent\textbf{Visual Activation Patching.} 
In an $L$-layer LVLM, the hidden state $s^l_i$ encodes token $x_i$ after being processed through layers $1$ to $l$. We first record hidden states from a forward pass on the corrupted image, then re-run inference on the clean image while replacing $s^l_i$ of a target token set at layer $l$ with its corrupted counterpart. Whether the model's prediction flips from the clean to the corrupted label gives the \textit{overwrite rate}, measuring the causal efficacy of the patched tokens.
We partition input tokens into six groups: System Prompt, Image Tokens, User Instruction, Generated Tokens, Last Image Token~(``|img\_end|''), and Last Prompt Token~(the ``Assistant:'' role tag), and use the prompt ``\textit{What is the number of the black dots in the image? Answer with the number only.}'' so that the first generated token is the counting answer.

\begin{figure}[htbp]
     \centering
  
     \begin{subfigure}[b]{0.51\textwidth}
         \centering
         \includegraphics[width=\textwidth]{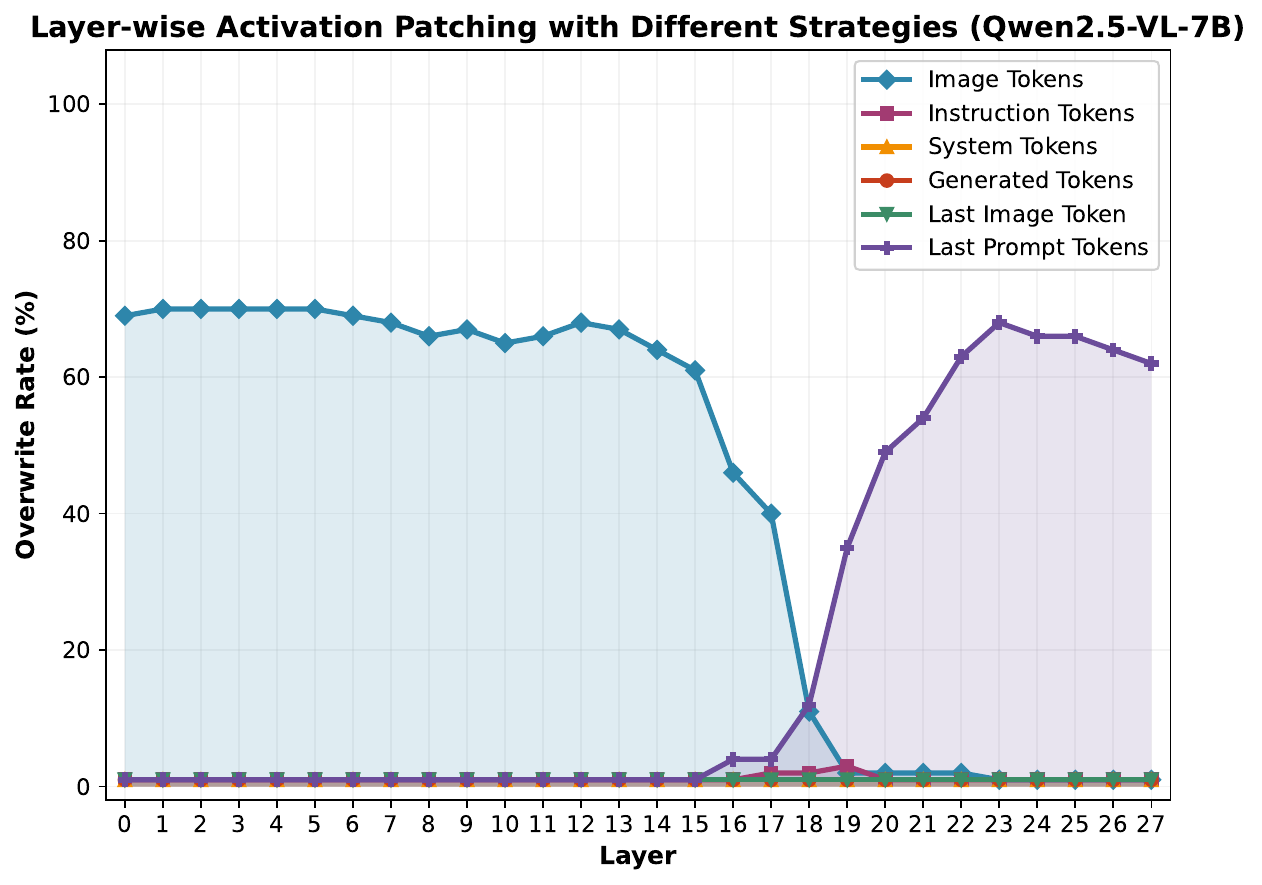}
         \caption{}
         \label{fig:qwen:overwrite}
     \end{subfigure}
     % \hfill 
     \begin{subfigure}[b]{0.48\textwidth}
         \centering
         \includegraphics[width=\textwidth]{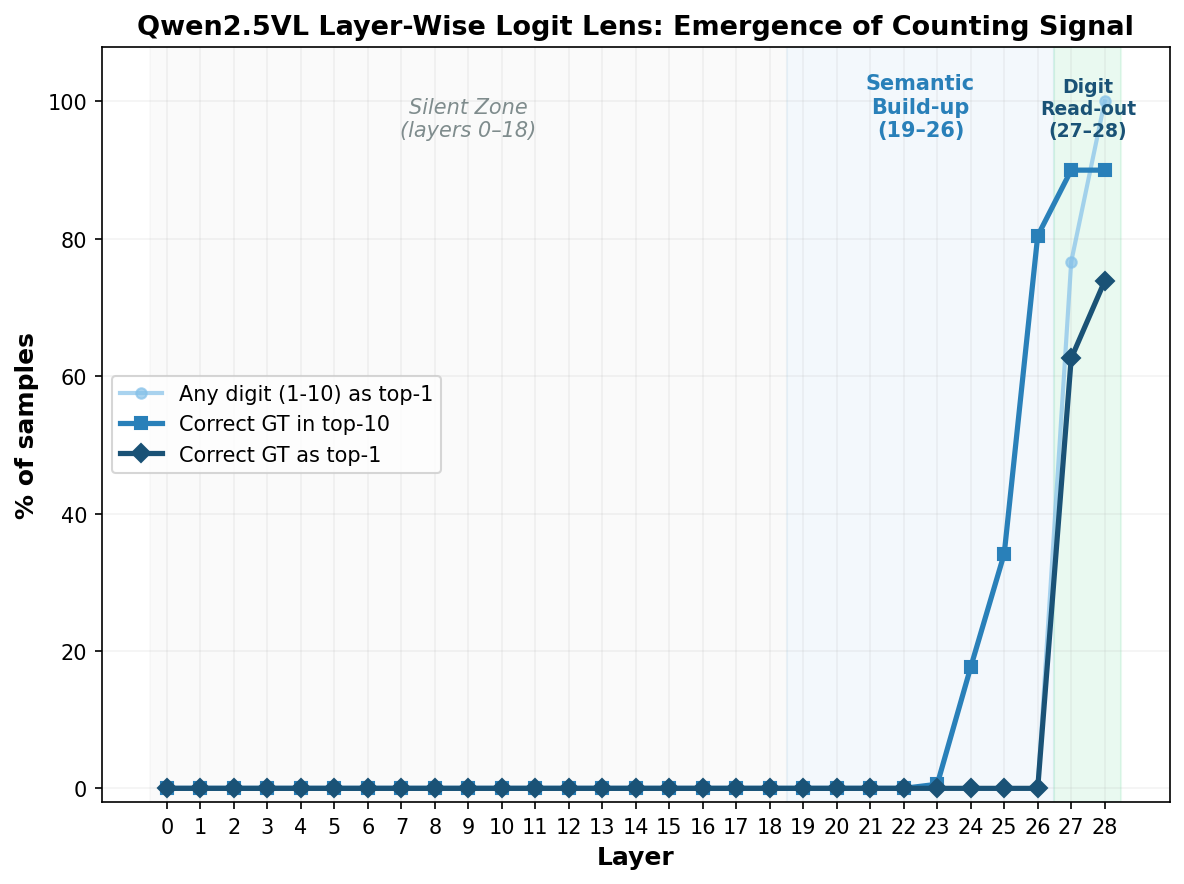}
         \caption{}
         \label{fig:qwen:layerwise_logit_lens}
     \end{subfigure}
     
     \caption{Left: Layer-wise Overwrite Rate of Different Input Tokens Patching Strategy; Right: Layer-wise Logit Lens Tracking Curve for Counting Number.}
     \label{fig:layerwise_analysis}
\end{figure}

% We conducted the layer-wise activation patching experiment with 100 black dot image pairs. As shown in~\cref{fig:qwen:overwrite}, in the early to middle layers ($L < 15$), the counting information is predominantly anchored within the \textit{All Image Tokens}, maintaining a high and stable causal influence. 
% However, starting from Layer $15$, the importance of image tokens drops significantly, while the causal efficacy of the \textit{Last Prompt Token} (the "Assistant:" role tag) surges, peaking around Layer $23$, showing a clear "handover" mechanism.
% This suggests that layers 15–22 serve as the critical bottleneck for cross-modal information routing.
% During this stage, spatially distributed visual features are compressed and "routed" into the linguistic stream.
% The negligible impact of user instruction or system prompt tokens further confirms that the numerical signal flows directly from visual representations to the response-triggering tokens. We observed similar cross-modal patterns for other LVLMs in the additional results in~\cref{sec:additional_interp}.

We run layer-wise activation patching on 100 black dot image pairs. As shown in~\cref{fig:qwen:overwrite}, in early-to-middle layers ($L < 15$), counting information is predominantly anchored in \textit{All Image Tokens} with high, stable causal influence.
From Layer $15$ onward, image token importance drops sharply while the \textit{Last Prompt Token} (``Assistant:'' tag) surges, peaking at Layer $23$, revealing a clear ``handover'' mechanism.
Layers 15--22 thus form the critical bottleneck for cross-modal routing, where spatially distributed visual features are compressed into the linguistic stream.
The negligible impact of user instruction and system prompt tokens confirms that the numerical signal flows directly from visual representations to the response-triggering tokens. Similar patterns hold for other LVLMs (\cref{sec:additional_interp}).

To corroborate that numerical representations emerge after the cross-modal routing phase, we employ a logit lens on the hidden state $s^l_{-1}$ immediately preceding the generation of the final counting token. 
By projecting these intermediate representations through the model's terminal unembedding layer, we map $s^l_{-1}$ directly into the vocabulary space to decode its latent semantic trajectory. 
As depicted in the~\cref{fig:qwen:layerwise_logit_lens}, correct number tokens begin to surface within the top-$10$ logit predictions between layers $19$ and $26$, gaining pronounced significance from layer $23$ onwards. 
By layers $27$ and $28$, the ground-truth counting token converges to the rank-1 position. 

This layer-wise progression captures the internal patterns of how the model aggregates visual numerosity, translates it into a discrete linguistic space, and ultimately solidifies the final numerical output. To further investigate the mechanistic underpinnings of these patterns, the subsequent section aims to pinpoint the specific attention heads responsible for executing the counting task and characterize their precise functionalities.

\section{Mechanistic Analysis on Attention Head Function}
\label{sec:head_analysis}

Isolating complete end-to-end circuits in LVLMs is intractable due to their massive scale and redundant backup heads~\cite{wang2022interpretability}. Therefore, we focus on identifying and revealing the critical attention heads that dominate the counting task.

\subsection{Important Attention Heads for Counting}

% \begin{figure}
%     \centering
%     \includegraphics[width=0.5\linewidth]{Figures/important_head.png}
%     \caption{Head-wise Importance Heatmap for Identifying Important Heads of Counting.}
%     \label{fig:important_head}
% \end{figure}

% \begin{wrapfigure}[15]{l}{0.4\textwidth}
%     \centering
%     \includegraphics[width=1\linewidth]{Figures/important_head.png}
%     \caption{Head-wise Importance Heatmap for Identifying Important Heads of Counting.}
%     \label{fig:important_head}
% \end{wrapfigure}

We conduct head-level visual activation patching to identify the important attention heads for counting tasks. Specifically, we leverage $100$ SynDot image pairs as the layerwise VAP but replace the target head activation. To prevent confounding effects from the residual stream, we define head activation strictly as the output of the projection layer. The importance score of each head is defined as the answer token logit difference change before and after the VAP. A positive importance score greater than $0.05$ indicates that the target head plays a significant role in the counting task. There are $\frac{43}{784} = 5.5\%$ heads noted as important heads. 
We visualize the heads with positive importance score as a heatmap in~\cref{fig:counting_heads} (left).
Next, we introduce a new interpretability tool named HeadLens and use it to reveal the functions of representative important heads based on their attention distribution and the semantic meaning of their output.

\subsection{HeadLens: Decoding Individual Attention Heads}

To isolate the semantic contribution of individual attention heads, we introduce a novel interpretability tool termed \textbf{HeadLens}. Utilizing the linear additivity of the multi-head self-attention (MHSA) projection, HeadLens enables granular, token-level semantic decoding without structural modifications.

Let $h_i(x) \in \mathbb{R}^{d_{\text{head}}}$ denote the output of the $i$-th attention head, where $H$ is the total number of heads and $d_{\text{model}} = H \times d_{\text{head}}$. We first isolate the $i$-th head's contribution by constructing a zero-padded representation $\tilde{h}_i(x) \in \mathbb{R}^{d_{\text{model}}}$:
\begin{equation}
    \tilde{h}_i(x) = [\underbrace{0, \ldots, 0}_{(i-1) \times d_{\text{head}}},\ h_i(x),\ \underbrace{0, \ldots, 0}_{(H-i) \times d_{\text{head}}}]
\end{equation}

Due to the properties of block matrices, the standard concatenated output of the MHSA mechanism can be equivalently formulated as the sum of these sparse representations. The final output of MHSA $\hat{x}$ is then obtained by applying the output projector matrix $W_O$:
\begin{equation}
    \hat{x} = \left( \sum_{i=1}^{H} \tilde{h}_i(x) \right) W_O = \sum_{i=1}^{H} \big( \tilde{h}_i(x) W_O \big)
\end{equation}

Rather than interpreting the aggregated hidden state $\hat{x}$, HeadLens operates directly on the isolated projection $\tilde{h}_i(x) W_O$. To translate this independent vector into human-interpretable concepts, we employ the learned affine translator~\cite{belrose2023eliciting} $T:\mathbb{R}^{d_{\text{model}}} \to \mathbb{R}^{d_{\text{model}}}$ (e.g., $T(z) = Az + b$) to map it into the final residual stream space.
Utilizing the model's unembedding matrix $U \in \mathbb{R}^{|\mathcal{V}| \times d_{\text{model}}}$ and bias $c$, we formulate the head-specific residual transformation $r_i(x)$, its corresponding logits $\ell_i(x)$, and the token probability distribution $p_i(\cdot \mid x)$ as follows:

\begin{equation*}
    r_i(x) = T \big( \tilde{h}_i(x) W_O \big) \qquad 
    \ell_i(x) = U\,r_i(x) + c \qquad 
    p_i(\cdot \mid x) = \mathrm{softmax} \big( \ell_i(x) \big)
\end{equation*}

By treating each $\tilde{h}_i(x) W_O$ as an independent semantic component, HeadLens effectively decodes the head activation into semantic tokens. We are particularly interested in tokens related to the visual features (e.g., color and shape) and counting (e.g., digits and numbers). We define the ratio of visual feature tokens and counting tokens in the top-10 HeadLens results as the Visual Grounding Score (VGS) and Counting Token Emergence Rate (CTER), respectively.

\subsection{Revealing Attention Head Functionalities}

In this section, we reveal the functionalities of four representative attention heads, including two counting heads (Cross-modal Routing Heads and Counting Aggregation Heads) and two special functional heads (Visual Grounding Heads and Awareness Heads) for the counting task.  
We provide the experiment details of head discovery in~\cref{sec:head_interpret_extend}.

\begin{figure}
    \centering
    \includegraphics[width=\linewidth]{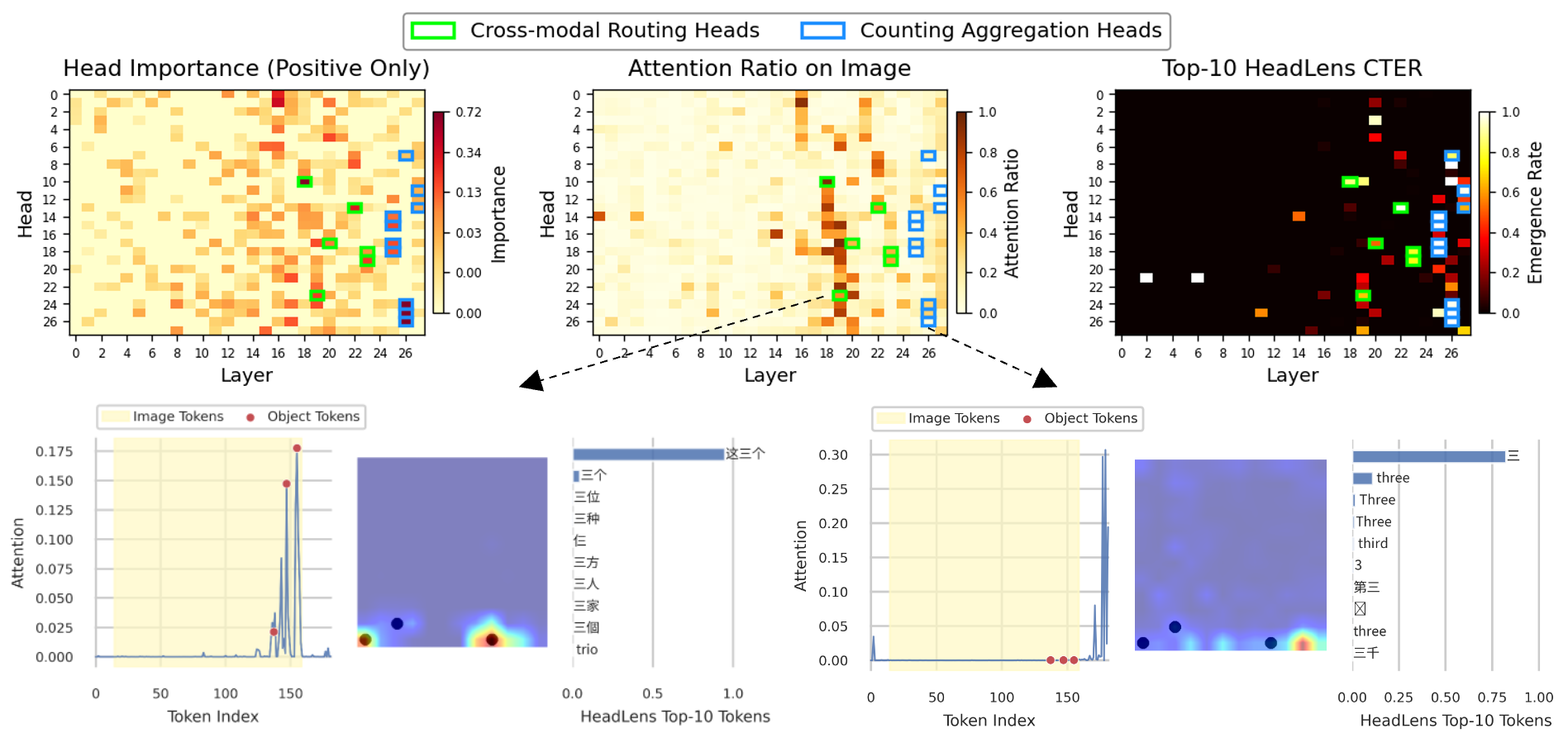}
    \caption{\textbf{Visualization of two functional groups of counting heads}. Left to right: Head importance (positive-only), attention ratio on image tokens, and top-10 HeadLens CTER across heads. We present the typical attention distribution and top-10 HeadLens results for Cross-modal Routing Heads~(green boxes) and Counting Aggregation Heads~(blue boxes), using L19H23(left) and L26H26(right). Both have "three"~({\begin{CJK}{UTF8}{gbsn}三\end{CJK}}) in Chinese as the top-1 HeadLens token.
    }
    \label{fig:counting_heads}
\end{figure}

% exhibit high image attention and high CTER, indicating their role in extracting object-related visual information and transferring it to number-relevant language tokens (e.g., L16H1). In contrast,  show low image attention but high head importance and CTER, suggesting they aggregate counting-relevant information from residual streams rather than directly attending to visual tokens (e.g., L26H26).

\noindent\textbf{Counting Heads.}
We discuss two functionally distinct attention head groups for the counting task, Cross-modal Routing Heads and Counting Aggregation Heads. 
\textbf{Cross-modal Routing Heads} are defined as heads with high attention ratio on image, high head importance, and high top-10 HeadLens CTER. These heads directly extract information from image tokens and transfer visual information into the language related to the counting task. As shown in \cref{fig:counting_heads}, these heads mainly lie in Layers 18–24 noted with green boxes. We visualize L19H23 as a typical example. It attends to object-related image regions and top-10 HeadLens tokens are relevant to number~(e.g., three), further confirming its role in bridging image and text. \textbf{Counting Aggregation Heads} are defined as heads with low image attention, high importance, and high top-10 HeadLens CTER. Despite extracting information from image tokens, these heads aggregate counting-relevant information already deposited in the residual stream by earlier layers rather than extracting visual information directly. We visualize L26H26, and its top-10 decoded token exhibit a remarkably high alignment with the model's final counting prediction while paying little attention to the image, confirming their role in counting aggregation.

\begin{figure}[t]
    \centering
    \includegraphics[width=1\linewidth]{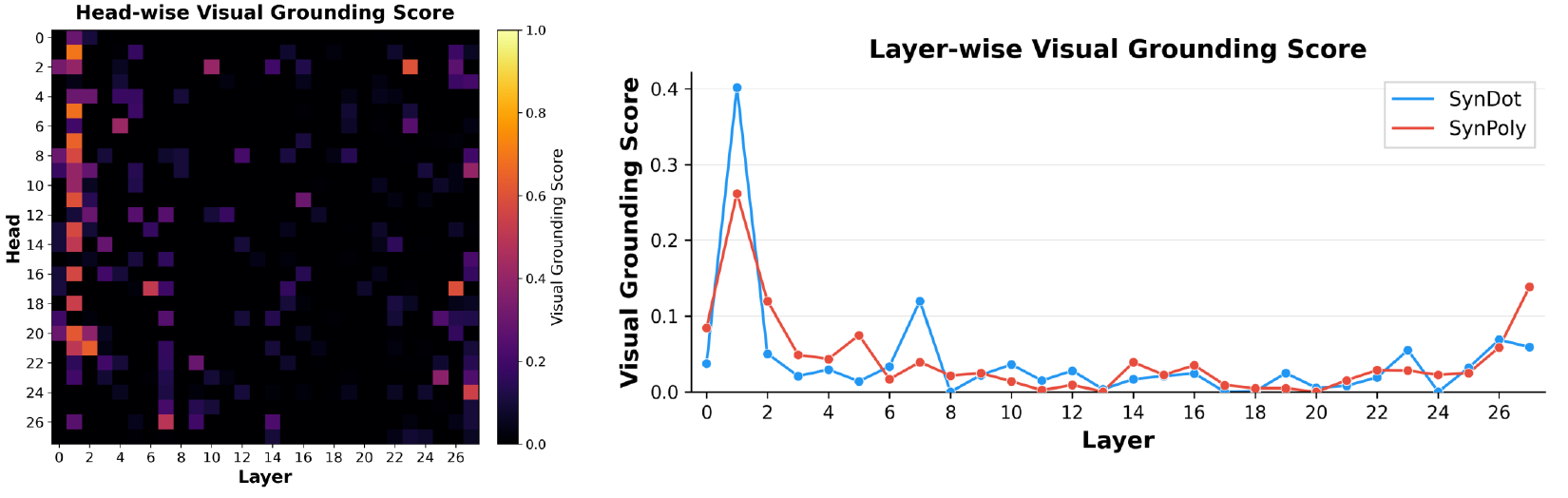}
    \caption{Visual Grounding Heads in Early Layers. Left: Head-wise heatmap based on Visual Grounding Score; Right: Layer-wise Visual Grounding Score.}
    \label{fig:visual_grounding_head}
\end{figure}

\noindent\textbf{Visual Grounding Heads.}
Concentrated primarily in the early stages of the network, specifically within Layer 1, these attention heads are dedicated to fundamental visual processing. They predominantly attend to object-specific image regions, serving as the primary mechanism for extracting basic, low-level visual attributes such as color, boundaries, and shape. We collect the HeadLens top-10 tokens of each head and compute the VGS accordingly. As shown in~\cref{fig:visual_grounding_head}, most visual grounding heads are located at layer 1, which has the highest average visual grounding score.

\noindent\textbf{Existence and Difficulty Awareness Heads.} We also identify two deep-layer attention heads that show awareness of object existence and the difficulty estimation of the counting.\textbf{L26H8} acts as a universal \textit{existence detector}, outputting ``1'' whenever any object is present. \textbf{L23H19} is a difficulty-aware head and one of the best counting aggregation heads ($32.9\%$ top-1 accuracy). Besides number tokens, its top-10 tokens are mixed with ``no'' and ``unnecessary'' for $GT=1$, indicating counting 1 dot is trivially easy. 
For $GT\geq2$, the head switches to digit predictions with reasonable accuracy, and for higher counts begins mixing in Chinese difficulty-related tokens like difficult and impossible, alongside the numeric predictions.

\section{Enhancing LVLMs' counting ability}
\label{sec:enhance_count}

% Our findings decode the step-by-step counting mechanism within LVLMs: early layers extract basic visual features via grounding heads, middle layers route visual information into a semantic latent space, and later layers aggregate the final counting answer. Guided by these structural characteristics, we introduce two targeted interventions to inherently boost the model's counting performance.

Our findings reveal a step-by-step counting mechanism: early layers extract visual features, middle layers route information into a semantic space, and later layers aggregate the answer. Guided by these findings, we introduce two targeted interventions.

\subsection{Object-Focused Attention Regularizer}

To improve the grounding of counting, we supervise the model's visual attention to be more ``object-centric'' using SynDot and SynPoly. Each image naturally provides object centers $\{c_k\}_{k=1}^{N}$, which we convert into a \textit{soft instance prior} on the $H \times W$ patch grid. We define an unnormalized grid score $u(p)$ at patch $p$ using a Gaussian kernel:$u(p)=\sum_{k=1}^{N}\exp\!\left(-\frac{\lVert p-\pi(c_k)\rVert_2^2}{2\sigma^2}\right)$,
where $\pi(\cdot)$ maps pixels to patch coordinates and $\sigma=1$ controls the supervision spread. We normalize $u(p)$ to obtain a target distribution $g \in \Delta^{|\mathcal{V}|-1}$ over the visual tokens $\mathcal{V}$.

We encourage the attention heads to allocate their weights on each image token $t \in \mathcal{T}$ to align with the instance prior $g$. For each layer $l$, we compute the average attention weights across all $H$ heads and renormalize it over $\mathcal{V}$ to obtain the predicted distribution $q^{\,l}_t$:
\begin{equation}
q^{\,l}_t(j) = \frac{\sum_{h=1}^{H} a_h^l(t,j)}{\sum_{j' \in \mathcal{V}} \sum_{h=1}^{H} a_h^l(t,j')}, \quad j\in\mathcal{V}.
\end{equation}
The focus loss is defined as the cross-entropy (equivalent to KL divergence) between $g$ and $q^{\,l}_t$, where $\epsilon$ is a small number for stability:
\begin{equation}
\mathcal{L}_{\text{focus}} = \frac{1}{|\mathcal{L}| |\mathcal{T}|} \sum_{l\in\mathcal{L}} \sum_{t\in\mathcal{T}} \left( -\sum_{j\in\mathcal{V}} g(j)\,\log\!\big(q^{\,l}_t(j)+\varepsilon\big) \right).
\end{equation}
Guided by our circuit discovery (\cref{sec:head_analysis}), we explore applying attention supervision to the visual grounding heads at early layers and cross-modal routing heads with a high image attention ratio at late layers to verify if the model's counting mechanism can be precisely localized and regularized.

\subsection{Adaptive Head Temperature Tuning}

Our analysis reveals that cross-modal routing and counting aggregation heads exhibit highly targeted attention patterns. To amplify this intrinsic circuitry, we introduce Adaptive Head Temperature Tuning, which reduces the attention entropy of these critical heads to sharpen their focus on relevant tokens. 

Instead of applying uniform temperature scaling, we adaptively modulate the pre-softmax logits of the identified target heads. For a given head $h$, we define an inverse temperature multiplier $\beta_h = \alpha \times \gamma_h$,
where $\alpha \ge 1$ provides a baseline entropy reduction, and $\gamma_h \ge 0$ is the head's intrinsic importance score derived from our circuit analysis. The modified attention matrix is thus computed as:
\begin{equation}
    A_h = \mathrm{softmax}\left( \beta_h \frac{Q_h K_h^T}{\sqrt{d_k}} \right)
\end{equation}

Applying this training-free technique exclusively to the routing and aggregation heads enhances the signal-to-noise ratio of the counting information flow.

\subsection{Joint Optimization Objective}
\label{sec:training_objective}

To integrate our targeted interventions, we combine standard Supervised Fine-Tuning (SFT) with our focus regularizer. The SFT process exclusively utilizes the SynDot and SynPoly datasets, restricting the autoregressive loss calculation strictly to the target counting tokens. The overall training objective is:
\begin{equation}
\mathcal{L} = \mathcal{L}_{\text{SFT}} + \lambda\,\mathcal{L}_{\text{focus}},
\label{eq:total_loss}
\end{equation}
where $\lambda$ controls the regularization strength (set to $1$ by default).

\section{Experiments}
\noindent\textbf{Model and Data.} We evaluate the proposed method on three different LVLMs: Qwen2.5-VL-7B, Qwen3-VL-8B and LLaVA-1.5-7B. The training data comprises $8000$ synthetic images equally from SynDot and SynPoly (classes 1--10, evenly per class). For OOD counting, we evaluate on SynReal and PixMo-Count~\cite{deitke2025molmo}.
For visual reasoning, we use MMMU~\cite{yue2024mmmu}, RealWorldQA~\cite{grok15v2024xai}, and MathVista~\cite{lu2023mathvista}.

\noindent{\textbf{Training.}}
We optimize with the joint loss (~\cref{eq:total_loss}). For SFT, we fine-tune models using LoRA ($r=64$) on the attention projection of all layers. We apply an attention regularizer to layers 2, 18-22 for Qwen2.5-VL-7B, layers 17-19 for Qwen3-VL-8B, and layers 0, 14 for LLaVA-1.5-7B, which have the most visual grounding heads and cross-modal routing heads.
All models are trained for $2$ epochs on a single NVIDIA H200 GPU. We use AdamW, BF16 precision, a batch size of $2$, and a learning rate of $2 \times 10^{-5}$ with linear decay and $3\%$ warmup. We take the mean results with 3 random seeds. For head tuning, we use a global value $\alpha = 1.2$ across different models.

\subsection{Evaluation Results}
\noindent\textbf{Counting Evaluation.} Our method consistently improves OOD counting across all three backbones on both SynReal and PixMo-Count. As shown in Table~\cref{tab:counting_eval_full}, the gains are consistent across all metrics, including higher accuracy and OBO, as well as lower MAE and RMSE. Notably, Qwen3-VL-8B, despite already being strong on SynReal, still shows clear gains, and LLaVA-1.5-7B also benefits substantially. These results show that our method transfers effectively to OOD counting rather than only fitting the training distribution. We provide the in-distribution (SynDot and SynPoly) and additional evaluation results in \cref{sec:additional_counting}.

\begin{table*}[ht]
\centering
\small
\setlength{\tabcolsep}{4pt}
\renewcommand{\arraystretch}{1.10}
\resizebox{\textwidth}{!}{%
\begin{tabular}{llcccc|cccc}
\toprule
\multirow{2}{*}{Backbone} & \multirow{2}{*}{Method}
& \multicolumn{4}{c|}{SynReal}
& \multicolumn{4}{c}{PixMo-Count} \\
\cmidrule(lr){3-6}\cmidrule(lr){7-10}
& & Acc & MAE$\downarrow$ & RMSE$\downarrow$ & OBO$\uparrow$ 
& Acc & MAE$\downarrow$ & RMSE$\downarrow$ & OBO$\uparrow$ \\
\midrule
\multirow{2}{*}{Qwen2.5-VL-7B}
& Baseline
& 73.48 & 0.79 & 2.13 & 91.01 
& 58.79 & 0.84 & 1.79 & 84.00 \\
& Ours
& \textbf{84.83} & \textbf{0.74} & \textbf{1.47} & \textbf{97.75} 
& \textbf{64.15} & \textbf{0.62} & \textbf{1.46} & \textbf{87.10} \\
\midrule

\multirow{2}{*}{Qwen3-VL-8B}
& Baseline
& 88.79 & 0.15 & 0.47 & 98.88 
& 58.75 & 0.73 & 1.35 & 88.20 \\
& Ours
& \textbf{91.21} & \textbf{0.11} & \textbf{0.34} & \textbf{99.41}
&  \textbf{66.98}  & \textbf{0.60} & \textbf{1.25} & \textbf{91.65}\\
\midrule

\multirow{2}{*}{LLaVA-1.5-7B}
& Baseline
& 54.27 & 3.59 & 5.05 & 66.34 
& 31.88 & 1.81 & 3.09 & 59.77 \\
& Ours
& \textbf{60.11} & \textbf{1.56} & \textbf{2.08} & \textbf{91.01} 
& \textbf{32.64} & \textbf{1.61} & \textbf{2.59} & \textbf{62.43} \\
\bottomrule
\end{tabular}%
}
\vspace{3pt}
\caption{Evaluation on Out-of-distribution Counting Benchmarks.}
\label{tab:counting_eval_full}
\end{table*}

\begin{table}[ht]
\centering
\small
\setlength{\tabcolsep}{6pt}
\renewcommand{\arraystretch}{1.08}
\begin{tabular}{llcccc}
\toprule
Backbone & Method & MMMU & RealWorldQA & MathVista & $\Delta$ (avg.) \\
\midrule
\multirow{2}{*}{Qwen2.5-VL-7B}
& Baseline & 54.89 & 61.96 & 57.30 & -- \\
& Ours & \textbf{56.33} & \textbf{64.14} & \textbf{58.30} & +1.54 \\
\midrule

\multirow{2}{*}{Qwen3-VL-8B}
& Baseline & 58.33 & 70.05 & 66.30 & -- \\
& Ours &  \textbf{60.74} & \textbf{71.83} & \textbf{67.90} & +1.93 \\
\midrule

\multirow{2}{*}{LLaVA-1.5-7B}
& Baseline & 44.44 & 56.21 & 23.90 & -- \\
& Ours & \textbf{44.56} & \textbf{56.48} & \textbf{26.30} & +0.93 \\
\bottomrule
\end{tabular}
\vspace{3pt}
\caption{Evaluation on General Capability Benchmarks.}
\label{tab:general_eval_clean}
\end{table}

\noindent\textbf{General Capabilities Evaluation.} Our method consistently improves general visual capabilities across three backbones, though trained on the counting task only. As shown in Table~\cref{tab:general_eval_clean}, all models achieve positive gains on MMMU, RealWorldQA, and MathVista, suggesting our intervention does not merely improve counting in isolation. Instead, it enhances the counting-related circuit which transfers to broader visual capabilities, indicating that counting may serve as a useful primitive for general visual reasoning. Notably, this improvement cannot be simply attributed to insufficient counting-related pretraining in the backbone. Qwen3-VL~\cite{bai2025qwen3vltechnicalreport} emphasizes stronger visual grounding and reasoning in its model and training design than Qwen2.5-VL~\cite{bai2025qwen2}, especially with additional counting training data.

\subsection{Ablation Study}

We decompose our full method into its individual components: LoRA-based SFT, Object-Focused Attention Regularizer ($\mathcal{L}_\text{focus}$), and Adaptive Head Temperature Tuning ($\beta_h$). As shown in~\cref{tab:ablation_main}, each component provides complementary gains. We provide additional ablation studies in~\cref{sec:additional_ablation}.

\begin{table}[ht]
\centering
\small
\setlength{\tabcolsep}{3pt} % 调整间距以适应 8 列数据
\renewcommand{\arraystretch}{1.08}
\caption{Component ablation on Qwen2.5-VL-7B across specialized and general benchmarks.}
\label{tab:ablation_main}
\begin{tabular}{ccc|ccccc}
\toprule
SFT & $\mathcal{L}_\text{focus}$ & $\beta_h$ & Synreal & PixMo-Count & MMMU & RealWorldQA & MathVista \\
\midrule
Baseline   &            &            & 73.48 & 58.79 & 54.89 & 61.96 & 57.30 \\
\checkmark &            &            & 79.78 & 61.67 & 56.00 & 63.14 &  58.10\\
\checkmark & \checkmark &            & 82.34 & 63.32 & 56.11 & 64.05 & 58.20 \\
\checkmark & \checkmark & \checkmark & \textbf{84.83} & \textbf{64.15} & \textbf{56.33} & \textbf{64.14} & \textbf{58.30} \\
\bottomrule
\end{tabular}
\end{table}

\subsection{Mechanistic Overlap: Why Does Counting Enhancement Generalize?}

% Intriguingly, we observed that enhancing the model's counting proficiency via our synthetic datasets (SynDot and SynPoly) yielded unexpected out-of-distribution (OOD) improvements on broader tasks, including general VQA and visual mathematical reasoning. To demystify the mechanistic origins of this transferability, we select five representative visual tasks for comparative analysis: (1) \textbf{Counting} (spanning synthetic SynDot and real-world CountBenchQA); (2) \textbf{Visual Attribution} (a custom diagnostic dataset querying the color and shape of a single synthetic polygon); (3) \textbf{Spatial Relations} (the CLEVR dataset); (4) \textbf{General VQA} (real-world QA); and (5) \textbf{Math Reasoning} (MathVista).

Enhancing counting performance via our synthetic datasets (SynDot and SynPoly) unexpectedly improved OOD performance on broader tasks, including general VQA and mathematical reasoning. To investigate the origins of this transferability, we analyze five representative visual tasks: (1) \textbf{Counting} (SynDot and Pixmo-Count), (2) \textbf{Visual Attribution} (a custom color and shape diagnostic), (3) \textbf{Spatial Relations} (CLEVR~\cite{johnson2017clevr}), (4) \textbf{General VQA}, and (5) \textbf{Math Reasoning} (MathVista).

\begin{figure}
    \centering
    \includegraphics[width=0.5\linewidth]{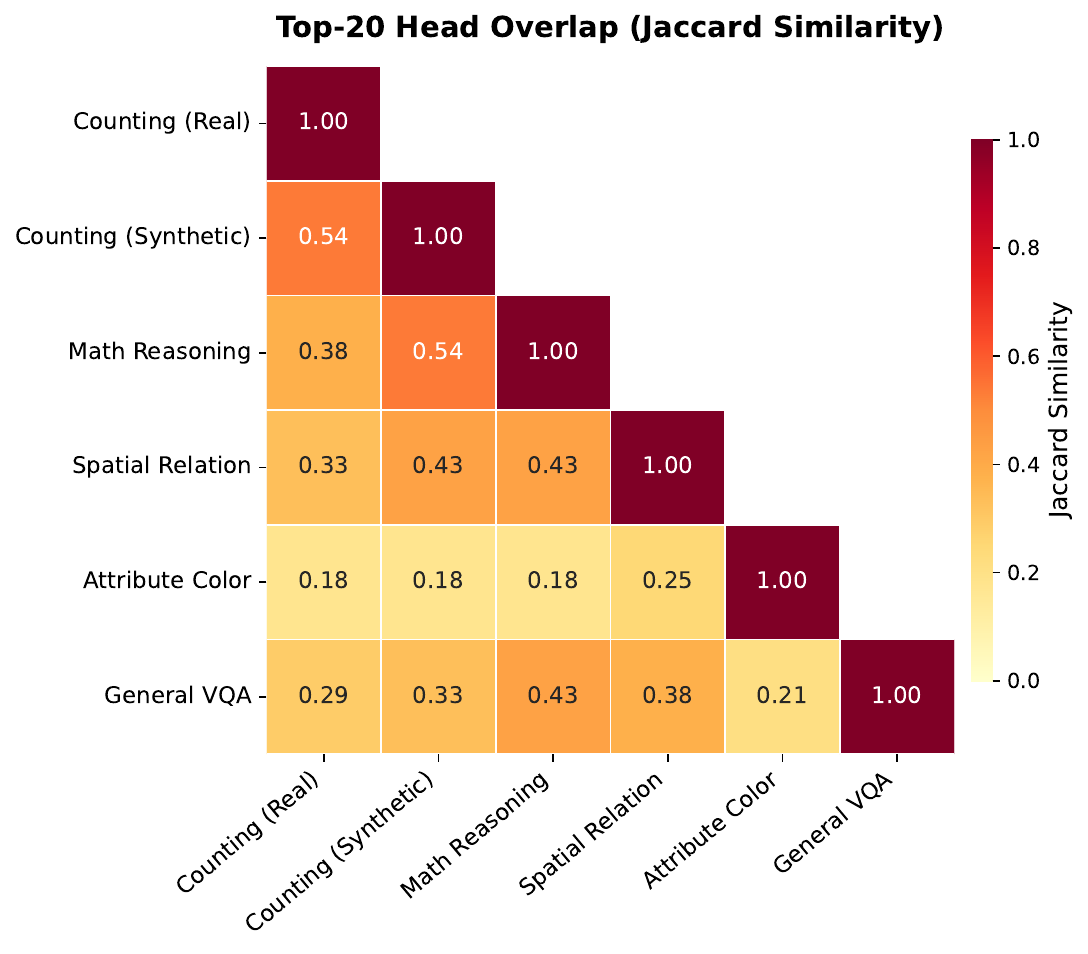}
    \caption{Jaccard Similarity for the top-20 most important heads of different tasks.}
    \label{fig:cross_task}
\end{figure}

% \vspace{-0.1in}
% \begin{wrapfigure}[15]{l}{0.46\textwidth}
%     \centering
%     \includegraphics[width=1\linewidth]{Figures/cross_task_similarity.pdf}
%     \caption{Jaccard Similarity for the top-20 most important heads of different tasks.}
%     \label{fig:cross_task}
% \end{wrapfigure}

To identify the underlying circuit mechanisms, we perform mean ablation at the head level. Specifically, we substitute each head's activation with its dataset-level mean and identify the Top-20 important heads based on the resulting logit degradation. We then use the Jaccard similarity $J$ to quantify the mechanistic overlap across tasks. We use $HS_A$ and $HS_B$ to denote two arbitrary important head sets.
\begin{equation}
    J(HS_A, HS_B) = \frac{|HS_A \cap HS_B|}{|HS_A \cup HS_B|} 
    \label{eq:jaccard}
\end{equation}

% As shown in \cref{fig:cross_task}, the \textit{Attribution} task exhibits minimal overlap with others, suggesting it relies on pure perception rather than reasoning. In contrast, the \textit{Counting} circuitry shows significant entanglement with reasoning-heavy tasks (sharing $>33\%$ of critical heads), explaining the observed OOD transferability. Notably, the high similarity between synthetic and real-world counting confirms that our enhancement targets fundamental enumeration circuits rather than dataset-specific artifacts. In~\cref{sec:color_shape_control}, we also verified that applying the same intervention method on the visual attribution task (asking the model to recognize the color and shape of colorful polygons) failed to improve the OOD visual reasoning performance of LVLMs. This further confirmed the fundamental role of the counting task.

As shown in \cref{fig:cross_task}, \textit{Attribution Color} exhibits minimal overlap with other tasks, indicating reliance on pure perception. In contrast, \textit{Counting} shares $>33\%$ of its critical heads with reasoning-heavy tasks, explaining the OOD transferability. The high similarity between synthetic and real-world counting further confirms that our method targets fundamental enumeration circuits rather than dataset-specific artifacts. In contrast, applying the same intervention to visual attribution (color/shape recognition) yields no OOD reasoning gains (\cref{sec:color_shape_control}), reinforcing counting's critical role.

% To identify the specific circuit mechanisms governing these tasks, we conduct a head-level activation patching analysis. Departing from standard paired-counterfactual patching, we perform \textit{mean ablation}: we substitute a specific head's activation with its dataset-level mean activation to evaluate its causal contribution to the final answer logit. For each task, we isolate the \textbf{Top-20 important heads} that induce the most severe logit degradation when ablated. We subsequently compute the \textbf{Jaccard similarity} between these head sets to quantify the mechanistic overlap, the shared circuit topologies across distinct tasks.

% \cref{fig:cross_task} reveals several profound insights. First, the \textit{Attribution} task exhibits remarkably low similarity with all other domains, revealing its nature as a pure perception task instead of a visual reasoning task like the others. Conversely, the \textit{Counting} circuitry demonstrates a high degree of entanglement with all reasoning-heavy tasks, sharing over a third of its critical heads with spatial, mathematical, and general VQA tasks. This structural congruence explains the OOD phenomenon as optimizing the counting-related heads and layers, also enhancing the shared sub-networks responsible for general visual reasoning. Furthermore, we observe the highest head-overlap rate between synthetic and real-world counting, validating that our synthetic enhancement strictly targets fundamental enumeration circuits rather than exploiting dataset-specific visual artifacts.
\section{Conclusion}

In this work, we show that counting is a meaningful probe of visual reasoning in LVLMs. Through controlled benchmarks and mechanistic analysis, we identify structured counting-related circuits and demonstrate that improving this primitive skill with lightweight synthetic-data intervention can yield gains beyond counting itself. These findings suggest that counting is not merely a narrow task, but an important building block of broader visual reasoning. Looking forward, an important direction is to examine whether these insights generalize to larger LVLMs and to other primitive visual skills, such as spatial reasoning and attribute comparison, toward a unified understanding of visual reasoning.

\bibliographystyle{IEEEtran}
\bibliography{ref}

\newpage
%%%%%%%%%%%%%%%%%%%%%%%%%%%%%%%%%%%%%%%%%%%%%%%%%%%%%%%%%%%%
\appendix

\section*{Appendix Table of Contents}
\startcontents[appendix]
\printcontents[appendix]{}{1}{\setcounter{tocdepth}{2}}
\newpage

\section{Preliminaries of Large Vision Language Models}
\label{sec:preliminaries}

A typical Large Vision Language Model (LVLM)~\cite{bai2025qwen2,liu2023visual,deitke2025molmo} consists of a vision encoder (e.g., vision transformer~\cite{dosovitskiy2020image}), a cross-modal alignment module (e.g., linear layer or MLP), and a Large Language Model (LLM) backbone. Given an input image $V$, the LVLM first splits it into $N_V$ image patches and then encodes them into a series of image tokens $[x_i^v]_{i=0}^{N_V}$. A paired instruction prompt $T$ will be tokenized and processed as text tokens $[x_i^t]_{i=0}^{N_T}$, where $N_T$ is the instruction token length. The final input sequences can also come with a system prompt claiming the LVLM's role in the conversation and tag prompt tokens such as "<image end>" and "Assistant:".

For a token $x_i$, its corresponding hidden states $s^{l}_{i} \in \mathcal{R}^{d_m}$ at the layer $l$ of the LLM backbone is updated through multi-head self-attention(MHSA) and feed-forward(FF) sublayers with residual connection. For the MHSA layer with $H$ heads, each head $h$ applies causal attention scores $Attention(Q,K,V) = softmax(\frac{QK^T}{\sqrt{d_h}}+M)V$ over the previous tokens. Here $Q,K,V\in \mathcal{R}^{n\times d_h}$ are the query, key, and value matrices projected from the input; $M$ is the causal mask. We denote $a^l_h$ as the attention scores of the head $h$ at layer $l$, and $a^l_h(i,j)$ is the attention score from token $x_i$ to token $x_j$.

\subsection{Architecture Comparison for Different LVLMs}

We discuss the architecture design difference between Qwen-2.5-VL, Qwen-3-VL and LLaVA-1.5 below. These three models have different designs on the vision encoder, cross-modal connector, and vision token construction methods.

% \noindent\textbf{InternVL3.5} The InternVL3.5-8B architecture follows a "ViT-MLP-LLM" design, employing an InternViT-300M as the vision encoder. To handle high-resolution inputs, it implements a Dynamic High Resolution strategy that tiles images of various aspect ratios into $448 \times 448$ patches. A key architectural component in this version is the Visual Resolution Router (ViR), a semantic-aware compression module. The ViR dynamically assigns different compression rates to image patches based on their complexity: complex patches undergo a standard $1/4$ compression, while simpler patches are subjected to a $1/16$ compression. This mechanism is designed to reduce the total number of visual tokens processed by the LLM while maintaining the integrity of essential visual information.

\noindent\textbf{Qwen2.5-VL.} Qwen2.5-VL-7B utilizes a native dynamic-resolution Vision Transformer (ViT), which is integrated with window-based attention mechanisms to ensure that computational complexity scales linearly with the number of input patches. For the vision-language connector, the model employs a spatially-aware pooling strategy that groups and concatenates sets of four adjacent patch features, followed by a two-layer MLP to condense the visual token sequence. Additionally, the architecture incorporates Multimodal Rotary Position Embedding (MROPE). This embedding is aligned to absolute time and spatial coordinates, allowing the model to process spatial scales and temporal sequences (for video tasks) within a unified positional framework without requiring traditional coordinate normalization.

\noindent\textbf{Qwen3-VL.} The Qwen3-VL 8B is a dense vision-language foundation model that integrates a SigLIP-2-SO-400M vision encoder with an 8B Qwen3 language model backbone to enable sophisticated multimodal reasoning and agentic decision-making. Architecturally, it employs a two-layer MLP-based merger to compress $2\times2$ visual features into single tokens while incorporating a pioneering DeepStack integration that injects multi-level ViT features directly into the first three hidden layers of the LLM via lightweight residual connections. To achieve robust spatial-temporal modeling, the framework utilizes an Interleaved MROPE scheme that uniformly distributes temporal, horizontal, and vertical dimensions across the frequency spectrum, alongside an explicit text-based timestamp strategy (e.g., "<3.0 seconds>") for precise video grounding. This design natively supports interleaved contexts of up to 256K tokens, allowing the 8B model to maintain strong pure-text proficiency while achieving performance competitive with much larger previous-generation models on complex long-document and video understanding tasks.

\noindent\textbf{LLaVA-1.5.} The LLaVA 1.5 architecture is designed as a direct bridge between a CLIP ViT-L/14 vision encoder and a large language model. The connection is established through a lightweight linear projection layer, which serves as the sole interface to map visual features $Z_v$ into the word embedding space of the language decoder. In this framework, the vision encoder functions as a visual tokenizer, converting images into a sequence of embeddings that the language model can process alongside text tokens. The model is trained end-to-end, focusing on the alignment between the visual features and the language model's latent space without the use of intermediate cross-attention modules or Q-formers.

\section{Datasets}
\label{sec:dataset}

\begin{figure}
    \centering
    \includegraphics[width=0.7\linewidth]{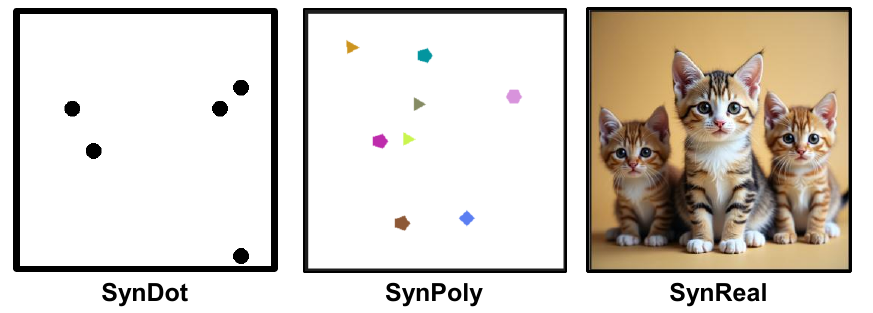}
    \caption{Data Samples of the three synthetic benchmarks.}
    \label{fig:syndata}
\end{figure}

\subsection{SynDot}
\label{data:syndot}
SynDot is a controlled synthetic dataset designed to isolate object counting from other visual complexities such as texture, color, and shape variation. Each image consists of a white canvas with a variable number of solid black circles (dots). Images are generated at a resolution of $336\times336$ pixels, where the canvas is partitioned into a grid of non-overlapping $28\times28$ patches. For each image, $N$ patches are randomly selected (where $N \in \{1, 2, \ldots, 10\}$), and a black circle of fixed radius (default 4 pixels) is drawn at the center of each selected patch. This grid-based placement guarantees that dots never overlap, allowing the ground-truth count to be unambiguously determined. This design eliminates confounding visual factors and enables a focused evaluation of a model's fundamental numerosity perception. The training set contains $4000$ images, and the test set comprises $2000$ images. 

% We additionally construct three diagnostic probing subsets from SynDot: (i)~\textit{Color}, pairing each black-dot image with a corresponding colorful-dot image sharing identical positions to test whether the model encodes color information; (ii)~\textit{Existence}, pairing dot-bearing images with blank white canvases to probe whether the model detects the presence of objects; and (iii)~\textit{Distribution}, contrasting uniformly scattered dots against spatially clustered dots to examine sensitivity to spatial arrangement.

\subsection{SynPoly}
\label{data:synpoly}
SynPoly extends the synthetic paradigm to incorporate greater visual diversity while remaining fully controlled. Each image is a $336\times336$ white canvas populated with $N$ randomly placed colorful polygons ($N \in \{1, 2, \ldots, 10\}$). For each polygon, the number of sides is uniformly sampled from $\{3, 4, 5, 6\}$ (i.e., triangles through hexagons), and the fill color is randomly drawn from the RGB space (excluding near-black and near-white values to ensure visibility). The polygon radius defaults to 8 pixels. By varying shape, color, and orientation simultaneously, SynPoly introduces within-image heterogeneity that more closely resembles real-world counting scenarios while still providing exact ground-truth labels. The test set contains $2000$ images ($200$ per count), and a larger training set of $4000$ images ($400$ per count) is prepared for supervised fine-tuning.

\subsection{SynReal}
\label{data:synreal}
SynReal bridges the gap between fully synthetic data and natural images by leveraging a state-of-the-art text-to-image diffusion model, FLUX.1-dev~\cite{labs2025flux1kontextflowmatching}, to generate photorealistic images containing specified quantities of real-world objects. We define six common object categories(\textit{cat}, \textit{dog}, \textit{bird}, \textit{car}, \textit{fish}, and \textit{person}) and generate images for each category with object counts ranging from 1 to 10. Each image is produced at $1024\times1024$ resolution. 
The text prompt follows the template ``\textit{a photo of \{N\} \{classname\}}''. 
For each category, $30$ images are generated with distinct random seeds, yielding an evaluation set of $180$ images.  
We conducted human verification for each of the generated images to ensure the label quality.
SynReal enables evaluation of counting ability under realistic visual conditions, including complex textures, occlusions, varying poses, and natural backgrounds.

\section{Counting Evaluation Settings}

\subsection{Evaluation Metrics}
\label{sec:eval_metric}

To comprehensively assess the visual counting capabilities of LVLMs, we employ a multi-dimensional evaluation suite. Relying solely on exact-match accuracy is insufficient for visual counting, as it fails to distinguish between minor estimation errors and catastrophic reasoning failures. Therefore, our metrics are designed to capture the exact precision, the magnitude of deviations, near-miss reliability, and the model's formatting compliance.

Let $N$ denote the total number of evaluated examples in a given dataset. For the $i$-th example, let $y_i \in \mathbb{N}$ represent the ground-truth object count, and $p_i$ denote the model's predicted count. To account for inference failures, where the model generates a vague, purely descriptive, or unparsable response instead of a discrete number, we assign a default penalty value of $p_i = -1$. We utilize the indicator function $\mathbf{1}[\cdot]$, which returns $1$ if the inner condition is true and $0$ otherwise. 

The four evaluation metrics are formally defined as follows:

\noindent\textbf{Accuracy (ACC).} 
Accuracy serves as the strictest measure of counting proficiency, calculating the proportion of predictions that exactly match the ground-truth label. 
$$ \mathrm{ACC} = \frac{1}{N} \sum_{i=1}^{N} \mathbf{1}[\, p_i = y_i \,] $$

\noindent\textbf{Mean Absolute Error (MAE).} 
Since visual counting is inherently a discrete regression task, MAE quantifies the average magnitude of numerical deviation from the ground truth. It provides an intuitive measure of how far off the model's predictions are on average, treating all linear errors equally.
$$ \mathrm{MAE} = \frac{1}{N} \sum_{i=1}^{N} \lvert p_i - y_i \rvert $$

\noindent\textbf{Root Mean Square Error (RMSE).} 
Unlike MAE, RMSE quadratically penalizes larger discrepancies before averaging. This metric is particularly sensitive to outliers or catastrophic counting failures (e.g., hallucinating a count of 10 when only 2 objects exist), making it a crucial indicator of the model's worst-case robustness.
$$ \mathrm{RMSE} = \sqrt{\frac{1}{N} \sum_{i=1}^{N} (p_i - y_i)^2} $$

\noindent\textbf{Off-by-one Accuracy (OBO).} 
In both human cognition (the Approximate Number System) and computer vision, minor counting errors ($\pm 1$) frequently occur due to occlusion, ambiguous boundaries, or subitizing thresholds. OBO measures the proportion of predictions that fall within a strict $\pm 1$ margin of error, serving as a proxy for "near-miss" reliability and functional estimation ability.
$$ \mathrm{OBO} = \frac{1}{N} \sum_{i=1}^{N} \mathbf{1}[\, \lvert p_i - y_i \rvert \le 1 \,] $$

\section{Additional Interpretability Results}
\label{sec:additional_interp}

\subsection{Revealing Attention Head Functionalities (Extended)}
\label{sec:head_interpret_extend}
In this section, we detail the experimental design and evaluation criteria for head function discovery. To systematically categorize the functional roles of individual attention heads, we evaluate them across three dimensions: importance scores, attention distribution patterns, and top-10 HeadLens projections.

\begin{figure}
    \centering
    \includegraphics[width=0.5\linewidth]{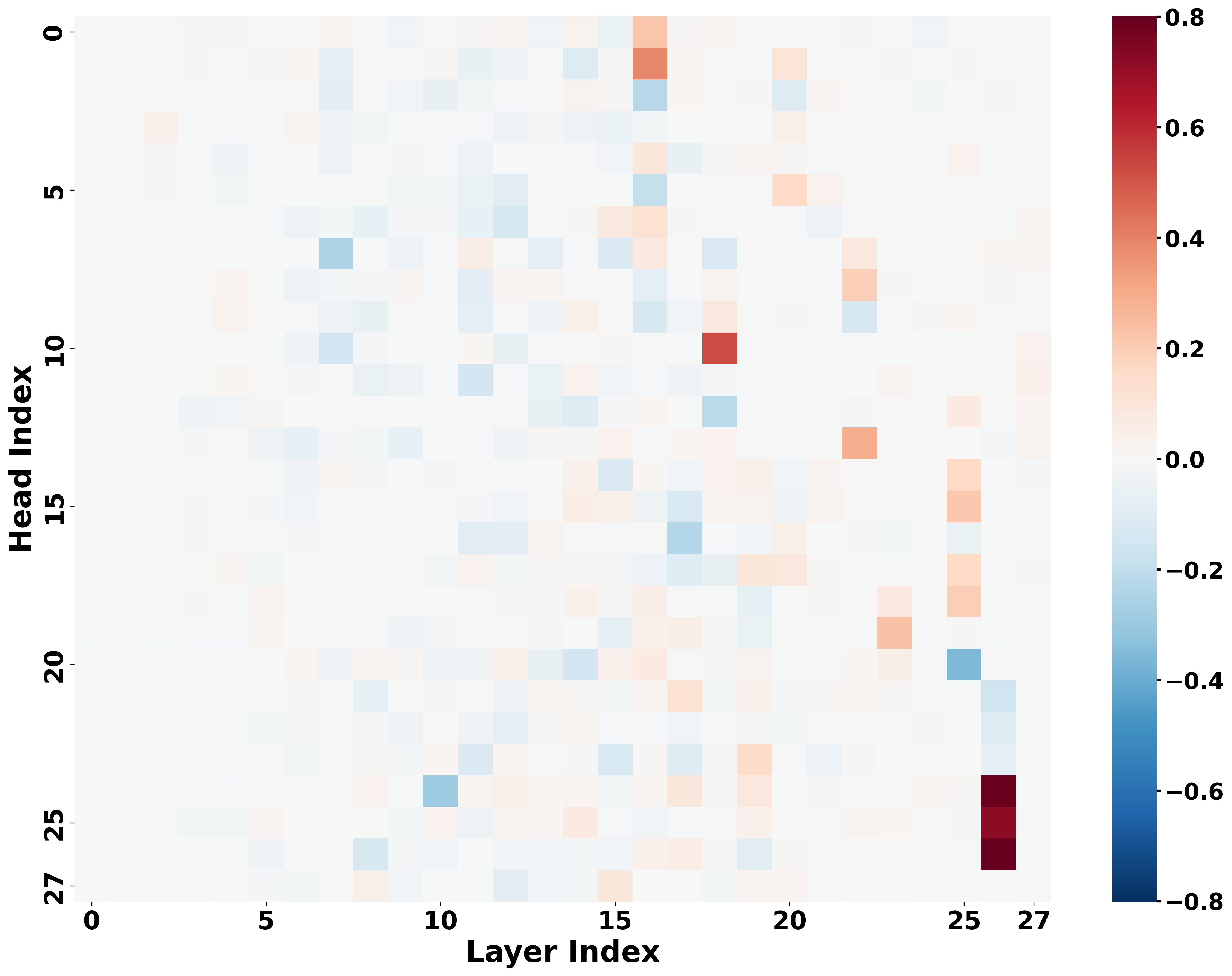}
    \caption{Head Importance Heatmap for Qwen2.5-VL on SynDot}
    \label{fig:importance_heatmap}
\end{figure}

Based on the overall distribution (visualized in \cref{fig:importance_heatmap}), we set an importance score threshold of 0.05, identifying 43 out of 784 heads (5.5\%) as functionally critical. 

For the attention distribution analysis, we first examine the allocation of weights between image and text tokens. Within the visual tokens, we further quantify the proportion of attention directed toward object-relevant regions. Consequently, our primary filtering step isolates heads that both exceed the $0.05$ importance threshold and allocate over $40\%$ of their total attention mass to image tokens. These heads are more likely to have visual-centric functions. We further evaluate the ratio of ground truth number tokens in top-10 HeadLens results and the top-1 HeadLens counting accuracy for each head. We categorize the identified influential heads into two functional groups based on their behavior across three metrics: (1) Counting Aggregation Heads (CAH), which exhibit high alignment with ground-truth counts in HeadLens projections ($Top\text{-}1\ Acc > 0.1$) despite lower direct image attention; and (2) Cross-Modal Routing Heads (CMR), which prioritize extracting spatial and object-relevant visual features ($Img\_Attn > 0.5$) but do not directly encode numerical information. We demonstrate the top-20 most important heads and their features in~\cref{tab:top20_heads}. Note that in practice, we combine the five metrics and their top-10 HeadLens frequency dictionary as the head categorization feature.

\begin{table}[t]
\centering
\caption{Top-20 attention heads ranked by importance and categorized by HeadLens projection. \textbf{Importance}: importance score from head ablation; \textbf{Img\_Attn}: fraction of attention on image tokens; \textbf{Obj\_in\_Img}: fraction of image attention on object-relevant regions; \textbf{GT@10}: ground-truth count in top-10 HeadLens projections; \textbf{Top-1 Acc}: top-1 HeadLens projection accuracy. Rows in \colorbox{cahgreen}{green} denote \textit{Counting Aggregation Heads} and rows in \colorbox{cmrblue}{blue} denote \textit{Cross-Modal Routing Heads}.}
\label{tab:top20_heads}
\resizebox{\linewidth}{!}{
\begin{tabular}{c l c c c c c}
\toprule
\textbf{Rank} & \textbf{ID} & \textbf{Import.} & \textbf{Img\_Attn} & \textbf{Obj\_in\_Img} & \textbf{GT@10} & \textbf{Top-1 Acc} \\
\midrule
\rowcolor{cahgreen}  1 & L26H26 & 0.801 & 0.034 & 0.032 & 0.83 & 0.55 \\
\rowcolor{cahgreen}  2 & L26H24 & 0.792 & 0.301 & 0.005 & 0.83 & 0.53 \\
\rowcolor{cahgreen}  3 & L26H25 & 0.716 & 0.314 & 0.023 & 0.92 & 0.77 \\
\rowcolor{cmrblue}   4 & L18H10 & 0.521 & 0.937 & 0.129 & 0.13 & 0.03 \\
\rowcolor{cmrblue}   5 & L16H1  & 0.393 & 0.903 & 0.260 & 0.10 & 0.00 \\
\rowcolor{cahgreen}  6 & L22H13 & 0.289 & 0.538 & 0.094 & 0.68 & 0.48 \\
\rowcolor{cahgreen}  7 & L23H19 & 0.232 & 0.470 & 0.033 & 0.32 & 0.17 \\
\rowcolor{cmrblue}   8 & L16H0  & 0.222 & 0.469 & 0.173 & 0.10 & 0.00 \\
\rowcolor{cmrblue}   9 & L22H8  & 0.198 & 0.708 & 0.024 & 0.07 & 0.00 \\
\rowcolor{cahgreen} 10 & L20H5  & 0.161 & 0.105 & 0.059 & 0.34 & 0.22 \\
\rowcolor{cahgreen} 11 & L25H17 & 0.157 & 0.155 & 0.041 & 0.69 & 0.30 \\
\rowcolor{cahgreen} 12 & L19H23 & 0.154 & 0.535 & 0.107 & 0.30 & 0.14 \\
\rowcolor{cahgreen} 13 & L20H1  & 0.104 & 0.015 & 0.083 & 0.21 & 0.10 \\
\rowcolor{cmrblue}  14 & L19H17 & 0.098 & 0.942 & 0.481 & 0.10 & 0.05 \\
\rowcolor{cmrblue}  15 & L19H24 & 0.087 & 0.689 & 0.014 & 0.10 & 0.00 \\
\rowcolor{cmrblue}  16 & L20H17 & 0.085 & 0.405 & 0.041 & 0.14 & 0.02 \\
\rowcolor{cmrblue}  17 & L22H7  & 0.084 & 0.564 & 0.059 & 0.10 & 0.00 \\
\rowcolor{cahgreen} 18 & L23H18 & 0.075 & 0.426 & 0.006 & 0.26 & 0.03 \\
\rowcolor{cmrblue}  19 & L16H18 & 0.064 & 0.515 & 0.029 & 0.11 & 0.00 \\
\rowcolor{cahgreen} 20 & L20H3  & 0.054 & 0.067 & 0.066 & 0.29 & 0.18 \\
\bottomrule
\end{tabular}
}
\end{table}

\subsection{Additional Layer-wise Visual Activation Patching Result}

We apply the same Layer-wise Visual Activation Patching experiment on LLaVA-1.5-7B, InternVL-3.5~\cite{wang2025internvl3} and Qwen3-VL-8B. It can be observed that both InternVL-3.5 and Qwen3-VL shows exactly the same cross-modal routing patterns as Qwen2.5-VL. The LLaVA-1.5 also shows the overwrite rate declining for the image tokens, indicating the counting information is transferred from visual information into the generated token directly. We hypothesize that LLaVA 1.5 fails to route counting information from the image tokens to the final prompt token because its corresponding circuitry remains underdeveloped during training, as illustrated in \cref{fig:qwen_llava_emergence}.

\begin{figure}[t]
    \centering
    \begin{subfigure}[t]{0.32\linewidth}
        \centering
        \includegraphics[width=\linewidth]{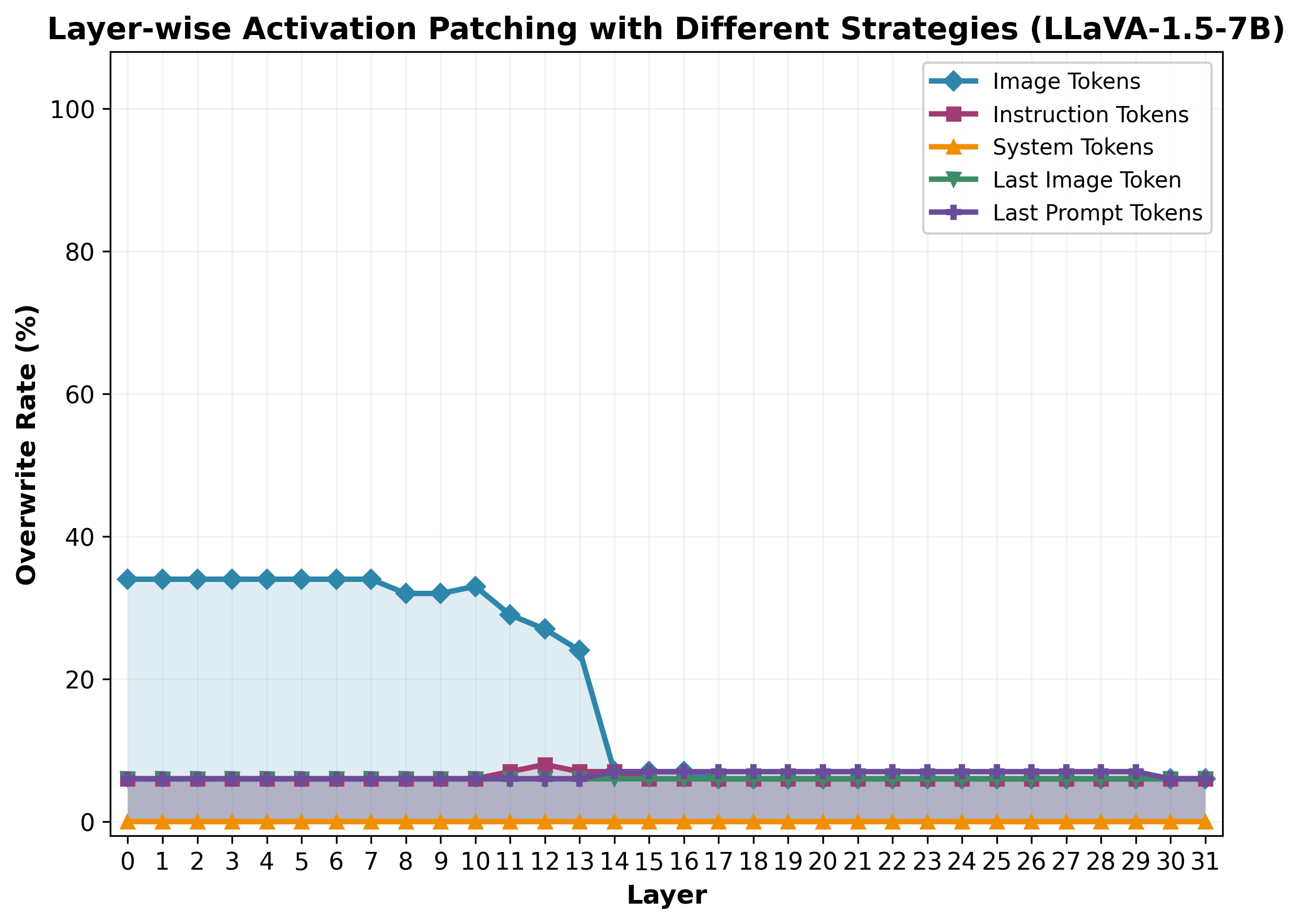}
        \caption{LLaVA-1.5-7B}
        \label{fig:VAP_llava}
    \end{subfigure}
    \hfill
    \begin{subfigure}[t]{0.32\linewidth}
        \centering
        \includegraphics[width=\linewidth]{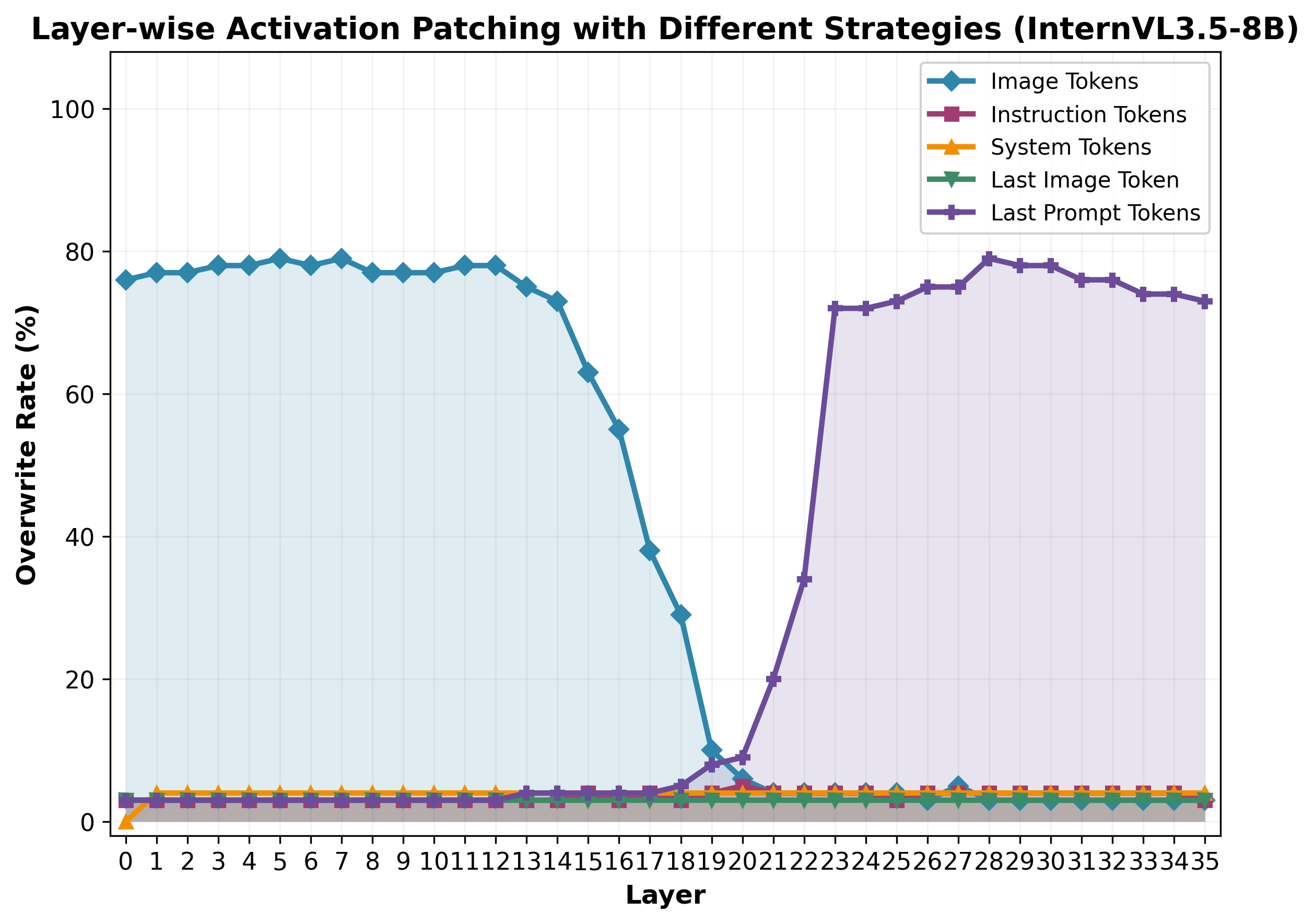}
        \caption{InternVL-3.5-8B}
        \label{fig:VAP_internvl}
    \end{subfigure}
    \hfill
    \begin{subfigure}[t]{0.32\linewidth}
        \centering
        \includegraphics[width=\linewidth]{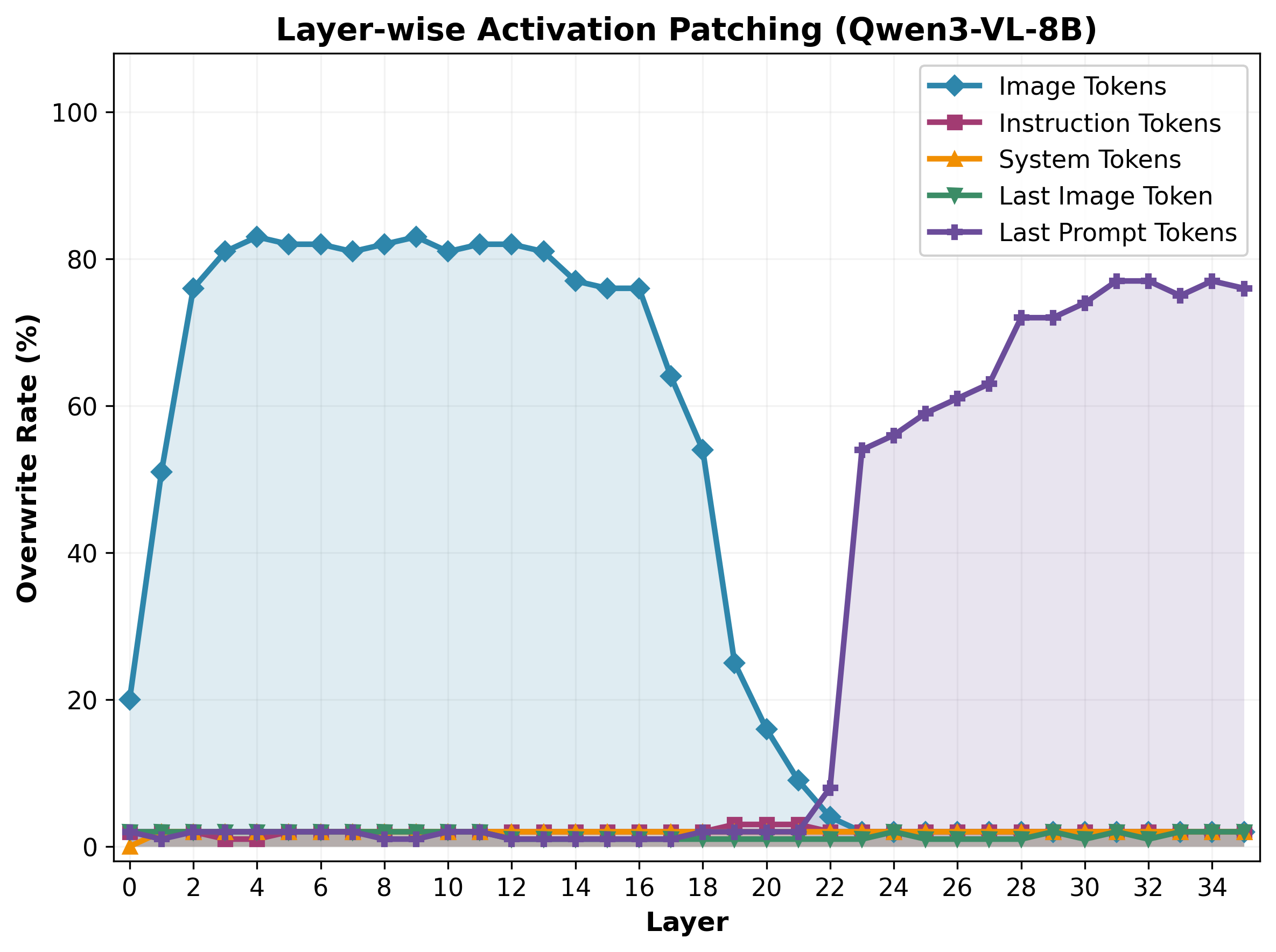}
        \caption{Qwen3-VL-8B}
        \label{fig:VAP_qwen3}
    \end{subfigure}
    \caption{Layer-wise VAP on SynDot for LLaVA-1.5-7B, InternVL3.5-8B and Qwen3-VL-8B.}
    \label{fig:vap_two_models}
\end{figure}

\subsection{Additional Head-wise Analysis}

\begin{figure}
    \centering
    \includegraphics[width=1\linewidth]{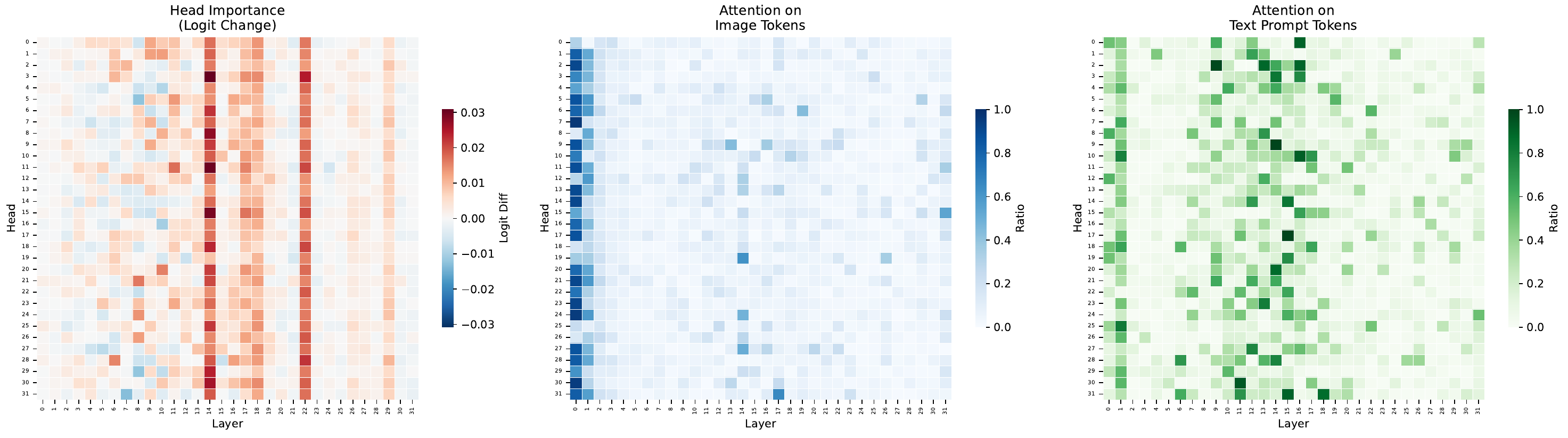}
    \caption{LLaVA1.5-7B: Head-wise HeatMap on Head Importance, Attention Ratio on Image Tokens and Text Tokens (from left to right).}
    \label{fig:llava_combined_importance}
\end{figure}

\begin{figure}
    \centering
    \includegraphics[width=1\linewidth]{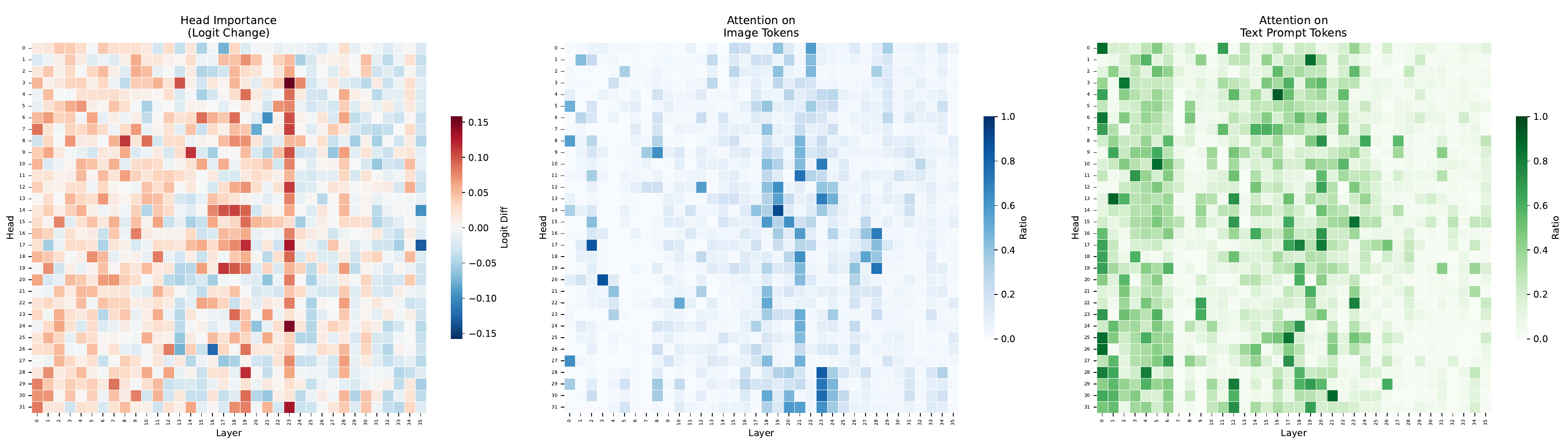}
    \caption{Qwen3-VL-8B: Head-wise HeatMap on Head Importance, Attention Ratio on Image Tokens and Text Tokens (from left to right).}
    \label{fig:qwen3_combined_importance}
\end{figure}

Similarly, we apply the head-level activation patching detailed in \cref{sec:head_analysis} to LLaVA1.5-7B and Qwen3-VL-8B to identify the attention heads critical for counting. As illustrated in \cref{fig:llava_combined_importance}, the most vital layers in LLaVA 1.5 are layers 14 and 22, while its first layer heads allocate the highest attention weights to image tokens. For Qwen3-VL (\cref{fig:qwen3_combined_importance}), the behavior mirrors Qwen2.5-VL: functionally important heads predominantly cluster in the middle layers and maintain the strongest focus on visual tokens.

\subsection{Probing the Dual-Stage Mechanism: Perception and Abstraction}
\label{sec:probe}

We posit that visual counting is not a monolithic process but comprises two distinct cognitive stages: {Perception}, which necessitates robust \textit{object binding}~\cite{treisman1998feature} to individuate distinct instances from the background; and {Reasoning}, which abstracts these visual signals into a  \textit{numerical quantity}. To empirically validate this hypothesis, we design two targeted probing tasks using the SynDot and CountBenchQA dataset~\cite{paiss2023teaching} for both synthetic and real-world objects.

\begin{figure}
    \centering
    \includegraphics[width=0.5\linewidth]{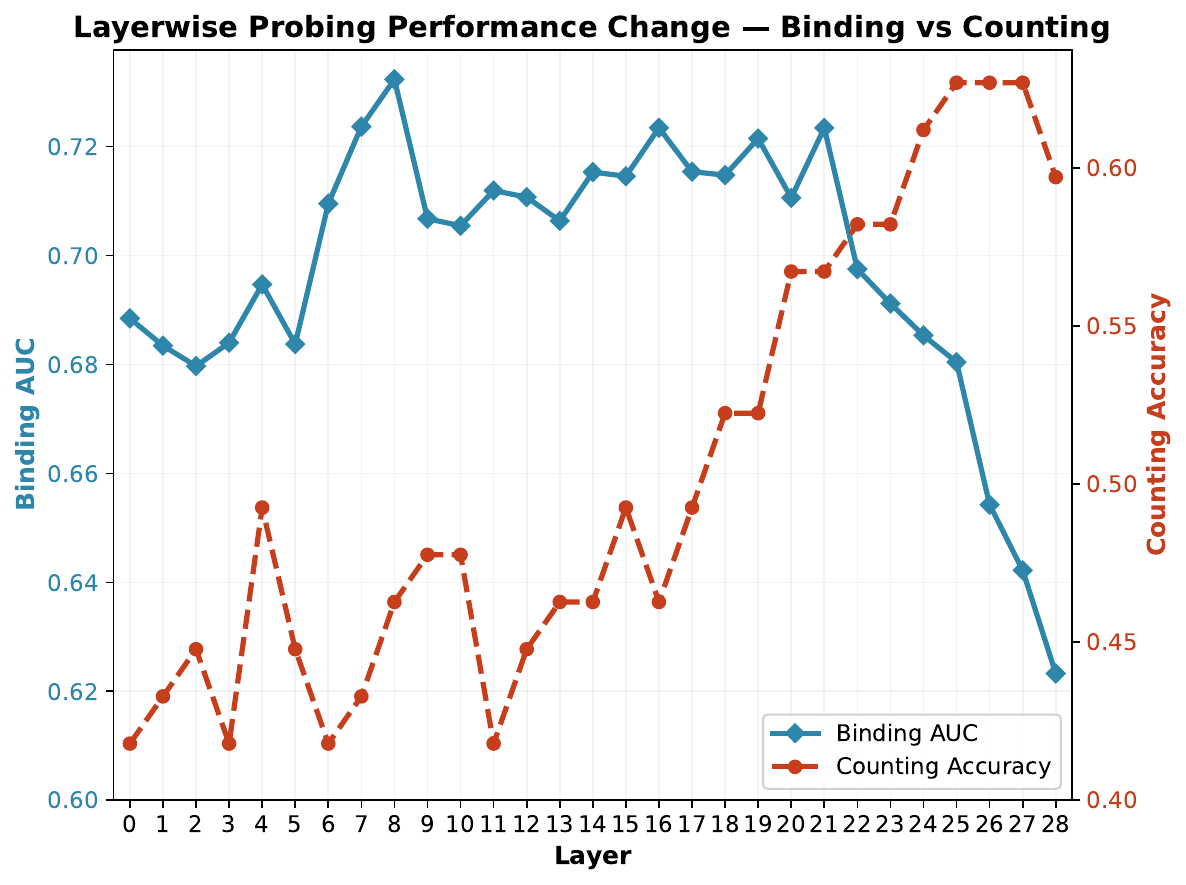}
    \caption{Layer-wise Probing Performance Curves for Binding and Counting.}
    \label{fig:binding_counting}
\end{figure}

% \begin{figure}[htbp]
%      \centering
  
%      \begin{subfigure}[b]{0.48\textwidth}
%          \centering
%          \includegraphics[width=\textwidth]{Figures/syndot_binding_counting.png}
%          \caption{Synthetic Objects(SynDot)}
%          \label{fig:left}
%      \end{subfigure}
%      % \hfill 
%      \begin{subfigure}[b]{0.48\textwidth}
%          \centering
%          \includegraphics[width=\textwidth]{Figures/countbench_binding_counting.pdf}
%          \caption{Real-world Objects(CountBenchQA)}
%          \label{fig:right}
%      \end{subfigure}
     
%      \caption{Layer-wise Probing Performance Curves for Binding and Counting.}
%      \label{fig:binding_counting}
% \end{figure}

\noindent\textbf{Data Curation Pipeline.} 
To isolate object-specific representations, we employ a two-stage preprocessing pipeline. First, {OwlViT-v2}~\cite{minderer2023scaling} performs open-vocabulary detection to localize bounding boxes for the target noun (e.g., "cats"). Subsequently, the SAM~\cite{kirillov2023segment} generates precise pixel-level masks for each detected box. This yields a set of instance-specific binary masks $\mathcal{M} = \{m_1, ..., m_N\}$ for each image, serving as the ground truth for our probes.

\noindent\textbf{Task A: Object Binding Probe (Perception).} 
This task evaluates whether the model's internal representations can distinguish between independent entities. Given two visual tokens $x^v_i$ and $x^v_j$, the probe predicts whether they belong to the same object instance. To capture relational subspace similarities that linear classifiers might miss, we follow~\cite{li2025does} to employ a Quadratic Probe with rank $r=64$. The similarity score is computed as:
\begin{equation}
    \text{Score}(x^v_i, x^v_j) = (W x^v_i)^\top (W x^v_j)
\end{equation}
where $W \in \mathbb{R}^{r \times d}$ is the learnable projection matrix. We measure performance using \textbf{ROC AUC}, quantifying the separability of instance-level features.

\noindent\textbf{Task B: Numerosity Probe (Abstraction).} 
This task assesses the model's ability to abstract a global count from visual features. We first compute the mean-pooled embedding of all tokens falling within the union of object masks, explicitly excluding background tokens to focus on object-centric representations. These aggregated embeddings are fed into an MLP classifier to predict the ground-truth count. Performance is evaluated via Accuracy, serving as a proxy for the distinctiveness of the model's internal numerical manifold.

As illustrated in \cref{fig:binding_counting}, the blue and red curves denote the layer-wise binding and counting performances, respectively. The binding AUC grows gradually through the early and middle layers before declining in the late layers as semantic information aggregates. Conversely, counting accuracy starts low but surges during the cross-modal routing stage between layers 15 and 23. This precise trajectory directly corroborates our layer-wise functional findings detailed in~\cref{sec:information_flow}.

\subsection{AttentionLens vs HeadLens}
\label{sec:attentionlens_vs_headlens}

% \begin{figure}
%     \centering
%     \includegraphics[width=1\linewidth]{Figures/heatmap_top1_acc_clean.png}
%     \caption{Head Heatmap of the accuracy between the head top-1 decoded token and the model's final output (not the ground truth counting). We compare the results between HeadLens and AttentionLens}
%     \label{fig:headlens_top1_align}
% \end{figure}

% \begin{figure}
%     \centering
%     \includegraphics[width=1\linewidth]{Figures/selectivity_kl_div_clean.png}
%     \caption{KL Divergence distribution for HeadLens and AttentionLens.}
%     \label{fig:headlens_kl_div}
% \end{figure}

In this section we provide a detailed comparison between HeadLens (ours) and AttentionLens~\cite{sakarvadia2023attention}, another head-level interpretability method. We first formalize both training pipelines, then argue that AttentionLens's unconstrained per-head probing introduces systematic noise that undermines circuit discovery, and finally show that HeadLens is orders of magnitude more efficient.

All experiments below are conducted on Qwen2.5-VL-7B-Instruct ($28$ layers $\times$ $28$ heads, $d_\text{head}\!=\!128$, $d_\text{model}\!=\!3584$, $|\mathcal{V}|\!=\!152{,}064$) using $2000$ SynPoly samples ($200$ per count class, classes 1--10).

\subsubsection{Training Formulations.}

\noindent\textbf{HeadLens} extends the Tuned Lens~\cite{belrose2023eliciting} to the per-head level.
\emph{Phase~1 (per-layer training):} For each layer $l$, a single affine translator $T_l({\mathbf{x}}) = W_l {\mathbf{x}} + {\mathbf{b}}_l$, $W_l \!\in\! \mathbb{R}^{d_\text{model} \times d_\text{model}},\, {\mathbf{b}}_l \!\in\! \mathbb{R}^{d_\text{model}}$, is trained to minimize:
\begin{equation}
    \mathcal{L}_{\text{HL}}^{(l)} = D_{\text{KL}}\!\Big(\,\mathrm{softmax}\!\big(\text{lm\_head}(\text{RMSNorm}(T_l({\mathbf{a}}_l)))\big) \;\Big\|\; \mathrm{softmax}({\mathbf{y}}_\text{final})\Big),
    \label{eq:headlens_loss}
\end{equation}
where ${\mathbf{a}}_l \!\in\! \mathbb{R}^{d_\text{model}}$ is the \emph{full} attention output at layer $l$ (post-$W_O$, pre-residual), and ${\mathbf{y}}_\text{final}$ is the model's final logits. All components except $(W_l, {\mathbf{b}}_l)$ are frozen; $W_l$ is initialized to identity and ${\mathbf{b}}_l$ to zero, ensuring the lens begins as a no-op.

\emph{Phase~2 (per-head decoding, no additional training):} To decode head $(l,h)$, we isolate its contribution via zero-padding and project through the corresponding $W_O$ slice:
\begin{equation}
    {\mathbf{p}}_{l,h} = W_O[:,\, h \cdot d_\text{head} \!:\! (h\!+\!1) \cdot d_\text{head}] \cdot {\mathbf{z}}_{l,h},
\end{equation}
where ${\mathbf{z}}_{l,h} \!\in\! \mathbb{R}^{d_\text{head}}$ is the raw head output. Then the trained translator and frozen unembedding produce:
\begin{equation}
    \text{logits}_{l,h} = \text{lm\_head}\!\big(\text{RMSNorm}(T_l({\mathbf{p}}_{l,h}))\big) \in \mathbb{R}^{|\mathcal{V}|}.
\end{equation}
HeadLens thus decodes every head through the model's \emph{own computational pathway}, reflecting each head's actual contribution to the residual stream.

\vspace{4pt}
\noindent\textbf{AttentionLens} trains a separate linear probe for \emph{every} head $(l,h)$:
\begin{equation}
    f_{l,h}({\mathbf{z}}_{l,h}) = W_{l,h}\, {\mathbf{z}}_{l,h} + {\mathbf{b}}_{l,h}, \quad W_{l,h} \!\in\! \mathbb{R}^{d_\text{model} \times d_\text{head}},\; {\mathbf{b}}_{l,h} \!\in\! \mathbb{R}^{d_\text{model}},
\end{equation}
trained to minimize:
\begin{equation}
    \mathcal{L}_{\text{AL}}^{(l,h)} = D_{\text{KL}}\!\Big(\,\mathrm{softmax}\!\big(\text{lm\_head}(\text{RMSNorm}(f_{l,h}({\mathbf{z}}_{l,h})))\big) \;\Big\|\; \mathrm{softmax}({\mathbf{y}}_\text{final})\Big).
    \label{eq:attentionlens_loss}
\end{equation}
Like HeadLens, AttentionLens reuses the model's frozen RMSNorm and lm\_head for decoding. However, the critical architectural difference is that it \emph{bypasses} the model's output projection $W_O$: instead of projecting head activations through the architectural $W_O$ slice, it learns an unconstrained linear map from $\mathbb{R}^{d_\text{head}}$ to $\mathbb{R}^{d_\text{model}}$. Each probe has ${\sim}$0.46M parameters (784 probes in total, ${\sim}$362M parameters overall) and is individually optimized per head.

\subsubsection{Critique: Forced Semantic Uniformity Undermines Circuit Discovery.}

The critical methodological concern with AttentionLens lies in \cref{eq:attentionlens_loss}: \emph{every} head is trained via KL divergence to approximate the model's final output distribution. This training objective implicitly assumes that each head's activation should be interpretable as a complete vocabulary distribution resembling the model's prediction. However, this assumption is fundamentally flawed:
\begin{itemize}
    \item \textbf{Not every head participates in the current task.} In a 784-head model, only a small fraction (${\sim}5.5\%$ as identified by activation patching in \cref{sec:head_interpret_extend}) causally contributes to counting. The remaining heads serve other functions (syntactic parsing, positional encoding, copy behavior) or are largely inactive for the given input.
    \item \textbf{Not every participating head is responsible for the final semantic output.} Even among task-relevant heads, many perform intermediate operations---e.g., cross-modal routing or visual grounding---whose internal representations are meaningful but do not directly resemble the final token distribution.
\end{itemize}

\begin{figure}[t]
    \centering
    % Top: side-by-side heatmaps
    \includegraphics[width=\linewidth]{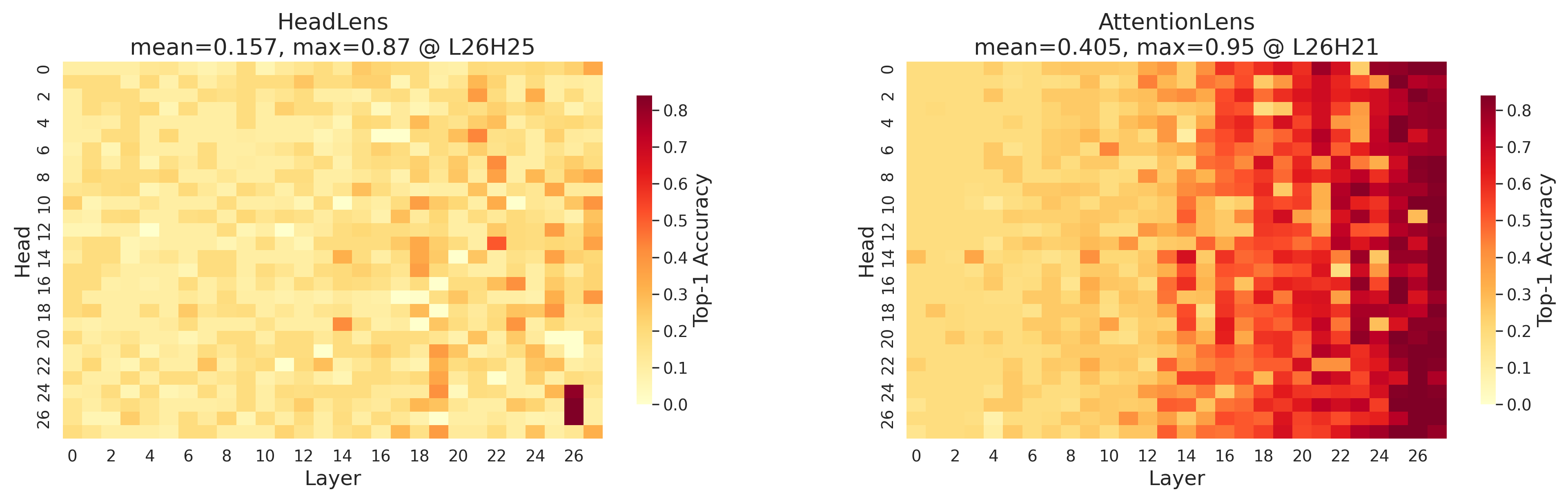}
    \\[4pt]
    % Bottom: KL distribution histogram
    \includegraphics[width=0.75\linewidth]{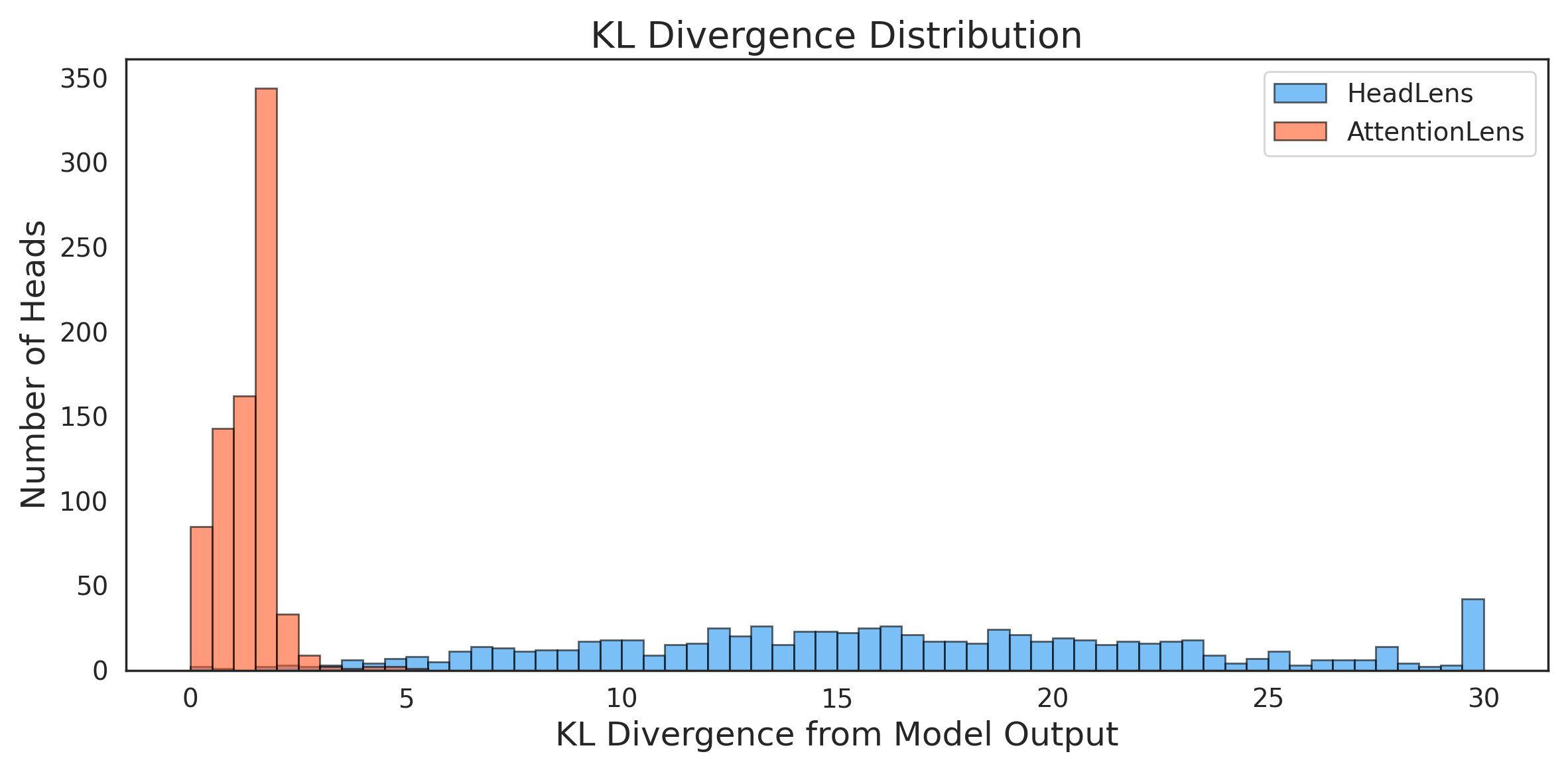}
    \caption{\textbf{Comparison of HeadLens and AttentionLens on SynPoly.} \textit{Top}: Head-wise top-1 accuracy on model final output (not ground truth counting). HeadLens (left) produces a sparse signal where only a few heads in Layer~26 stand out, which is aligned with our head importance heatmap in~\cref{fig:importance_heatmap}. While AttentionLens (right) spreads high accuracy across the majority of heads, especially in layers 14--27. \textit{Bottom}: KL divergence distribution across all 784 heads. AttentionLens concentrates nearly all heads into a narrow low-KL band ($\text{KL} < 3$), collapsing the dynamic range needed to distinguish critical from non-critical heads. HeadLens preserves a broad distribution that naturally separates the counting circuit from irrelevant heads.}
    \label{fig:hl_al_comparison}
\end{figure}

Despite this, the KL-divergence training in \cref{eq:attentionlens_loss} \emph{forces} each probe to fit the final distribution regardless. With no architectural constraint tying the learned map to $W_O$, each probe has full freedom to discover whatever linear relationship between $\mathbb{R}^{d_\text{head}}$ and $\mathbb{R}^{d_\text{model}}$ best approximates the target, even for heads whose actual $W_O$ projection contributes nothing task-relevant. The result, as shown in \cref{fig:hl_al_comparison}, is that AttentionLens makes nearly \emph{every} head appear to produce a distribution closely aligned with the model's final output.

This ``semantic flattening'' effect is directly visible in \cref{fig:hl_al_comparison}. Under AttentionLens, the top-1 accuracy on model response heatmap shows a large, densely activated region spanning most of layers 14--27, with a mean top-1 accuracy of $0.405$ (compared to HeadLens's 0.157). The KL divergence histogram further reveals the problem: AttentionLens compresses virtually all 784 heads into a narrow low-KL band ($\text{KL} < 3$), while HeadLens preserves a broad spread from 0 to 30. Under AttentionLens, the few genuinely important counting heads (L26H24--26) are buried within hundreds of heads that \emph{appear} equally capable.

For mechanistic interpretability, this is counterproductive. The goal of circuit discovery is to identify the \emph{minimal} set of attention heads that causally drive a behavior, which requires a method that clearly separates contributing heads from non-contributing ones.
To quantify this, we compute the best-to-mean VGS ratio as a discriminability measure: HeadLens yields $0.772 / 0.134 = 5.8\times$, while AttentionLens yields only $0.639 / 0.229 = 2.8\times$. The sparser HeadLens signal makes it straightforward to threshold and identify the counting circuit; the flatter AttentionLens signal requires additional filtering that reintroduces subjectivity.

\subsubsection{HeadLens: Faithfulness via Architectural Consistency.}

HeadLens avoids the above pitfall because its translator is trained on the \emph{full aggregate} attention output ${\mathbf{a}}_l$ (\cref{eq:headlens_loss}) rather than on individual heads. The translator learns to bridge the representational gap between intermediate attention outputs and the final layer, a correction that is \emph{shared} across all heads within a layer. When this shared translator is subsequently applied to a \emph{single} head's projected output ${\mathbf{p}}_{l,h}$, it does not force that head to mimic the final distribution. Instead, heads that genuinely contribute counting-relevant information through $W_O$ produce strong decoded signals, while heads that contribute little or encode information in subspaces projected away by $W_O$ naturally produce weak or incoherent signals.

This property follows from the causal structure of the transformer itself: the residual stream at each layer is the sum of all head contributions, and the final output is causally determined by these accumulated contributions passed through RMSNorm and lm\_head. By preserving this pathway, HeadLens measures what each head actually \emph{writes into} the residual stream rather than what can be \emph{extracted from} its activation by an unconstrained probe. The distinction is precisely the difference between \emph{causal contribution} and \emph{information existence}---the former is what circuit discovery requires.

\subsubsection{Computational Cost.}

Beyond the methodological concerns, AttentionLens incurs substantial computational overhead due to the sheer number of independently trained probes (\cref{tab:hl_al_cost}). Although both methods reuse the model's frozen lm\_head and have comparable total parameter counts (${\sim}$360M), AttentionLens must train 784 independent probes (each $\mathbb{R}^{d_\text{head}} \!\to\! \mathbb{R}^{d_\text{model}}$) with ${\sim}$235K total optimization steps, whereas HeadLens trains only 28 shared translators (each $\mathbb{R}^{d_\text{model}} \!\to\! \mathbb{R}^{d_\text{model}}$) with ${\sim}$14K steps.

\begin{table}[h]
\centering
\small
\setlength{\tabcolsep}{5pt}
\renewcommand{\arraystretch}{1.08}
\caption{Computational cost comparison between HeadLens and AttentionLens.}
\label{tab:hl_al_cost}
\begin{tabular}{lccc}
\toprule
 & HeadLens & AttentionLens & Ratio \\
\midrule
\# Lenses trained       & 28          & 784           & $28\times$ \\
Parameters per lens     & ${\sim}$12.8M       & ${\sim}$0.46M         & $0.04\times$ \\
Total parameters        & ${\sim}$360M        & ${\sim}$362M          & ${\sim}1\times$ \\
Training steps          & 14{,}000    & 235{,}200     & $17\times$ \\
Wall time (SynPoly)     & 37.9\,s     & 2{,}385.1\,s  & $63\times$ \\
Wall time (SynDot)   & 41.9\,s     & 2{,}414.5\,s  & $58\times$ \\
\bottomrule
\end{tabular}
\end{table}

In practice, HeadLens completes full-model analysis in under one minute, whereas AttentionLens requires ${\sim}$40 minutes on the same hardware (NVIDIA RTX A6000)---a $\mathbf{{\sim}60\times}$ speedup. Despite comparable total parameter counts, the $28\times$ difference in the number of independently trained lenses dominates the wall-clock cost due to repeated optimizer initialization, data loading, and per-probe forward/backward passes.
This efficiency gap is especially consequential for iterative workflows in mechanistic interpretability, where practitioners repeatedly analyze head behaviors across different inputs, tasks, and model checkpoints.

\subsubsection{Summary.}
Both methods localize the same core counting circuit (Layer~26, Heads~24--26), confirming the robustness of this finding. However, HeadLens is strongly preferred for circuit discovery:
\begin{enumerate}
    \item \textbf{Faithful decoding}: HeadLens measures each head's causal contribution through the model's own pathway ($W_O \!\to\! T_l \!\to\! \text{RMSNorm} \!\to\! \text{lm\_head}$), whereas AttentionLens bypasses $W_O$ with a freely learned linear map, conflating information \emph{existence} with information \emph{usage}.
    \item \textbf{Discriminability}: By not forcing every head to mimic the final output, HeadLens naturally produces a sparse signal (best/mean VGS ratio $5.8\times$ vs.\ $2.8\times$) that cleanly separates counting-critical heads from irrelevant ones. AttentionLens's forced semantic uniformity makes nearly all heads appear meaningful, introducing noise that hinders both circuit discovery and functional decomposition.
    \item \textbf{Efficiency}: Despite comparable total parameter counts, HeadLens runs ${\sim}60\times$ faster by training only 28 shared translators instead of 784 independent probes, making it practical for routine, large-scale head-level analysis.
\end{enumerate}

\subsection{HeadLens Results Comparison Between LLaVA1.5-7B and Qwen2.5-VL-7B}

We compare the architecture difference between LLaVA1.5-7B and Qwen2.5-VL-7B based on the HeadLens discovery based on SynDot dataset. 

% \noindent\textbf{Top Counting Head Comparison.}
% As shown in~\cref{fig:qwen_llava_tophead}, the most striking architectural difference between the two models is the distribution of counting ability across attention heads. LLaVA concentrates its counting ability in essentially one head (L29H5, $30.9\%$ correct-in-top-10), with a steep dropoff to the next head ($19.9\%$) and rapidly diminishing returns. Qwen2.5VL, by contrast, distributes the counting signal across a broad coalition of heads: five heads each achieve $80\%$ correct-in-top-10 (L25H14, L25H15, L25H18, L27H11, L27H13), and an additional 9 heads exceed $17\%$. This distributed architecture provides redundancy and allows different heads to specialize in different aspects of the counting task.

% \begin{figure}
%     \centering
%     \includegraphics[width=\linewidth]{Figures/qwen_llava_tophead.png}
%     \caption{Top Counting Head Performance Comparison between Qwen2.5-VL 7B and LLaVA1.5-7B}
%     \label{fig:qwen_llava_tophead}
% \end{figure}

\begin{figure}
    \centering
    \includegraphics[width=\linewidth]{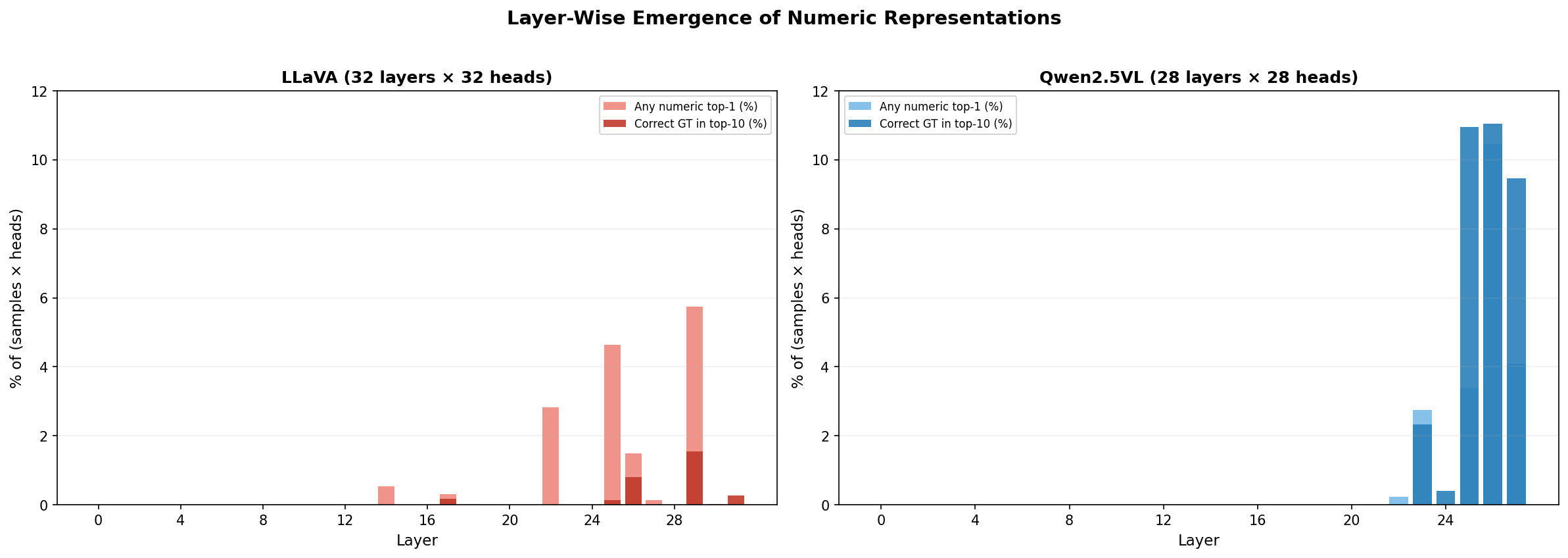}
    \caption{Layer-wise emergence of numeric representations for both models. Both show zero numeric signal in the first ~$70\%$ of layers. Qwen's emergence is more concentrated (layers 25–27) and achieves far higher correct-in-top-10 rates (up to $11\%$) versus LLaVA's peak of ~$1.6\%$.}
    \label{fig:qwen_llava_emergence}
\end{figure}

\noindent\textbf{Layer-Wise Emergence Comparison.}
As in~\cref{fig:qwen_llava_emergence}, both models share the pattern of late emergence-numeric representations appear only in the final ~$20\%$ of the network. However, Qwen's emergence is more concentrated and more powerful. While LLaVA shows scattered, weak numeric signals across layers 14–31, Qwen is completely silent until layer 22 and then explodes with a strong signal across layers 23–27. Qwen's peak per-layer correct-in-top-10 rate ($11\%$) is $7\times$ higher than LLaVA's peak ($1.6\%$), consistent with its distributed counting architecture. Qwen compresses its counting computation into a narrower, deeper band of layers ($79-96\%$ depth) but deploys far more heads within that band. LLaVA spreads weak signals over a wider layer range ($44-97\%$ depth) but never achieves meaningful concentration. This suggests Qwen's architecture enables more efficient "late-stage" counting circuits.

\section{Additional Experiment}

\subsection{In Distribution Counting Evaluation}
\label{sec:in_distribution_counting}
\begin{table*}[h]
\centering
\small
\setlength{\tabcolsep}{4pt}
\renewcommand{\arraystretch}{1.10}
\resizebox{\textwidth}{!}{%
\begin{tabular}{llcccc|cccc}
\toprule
\multirow{2}{*}{Model} & \multirow{2}{*}{Method} & \multicolumn{4}{c|}{SynDot} & \multicolumn{4}{c}{SynPoly} \\
\cmidrule(lr){3-6}\cmidrule(lr){7-10}
& & Acc & MAE$\downarrow$ & RMSE$\downarrow$ & OBO$\uparrow$ & Acc & MAE$\downarrow$ & RMSE$\downarrow$ & OBO$\uparrow$ \\
\midrule
\multirow{2}{*}{Qwen2.5-VL-7B} & Baseline & 76.12 & 0.25 & 0.52 & 96.21 & 53.74 & 0.94 & 1.75 & 82.27 \\
& Ours & \textbf{89.20} & \textbf{0.11} & \textbf{0.33} & \textbf{100.00} & \textbf{79.10} & \textbf{0.21} & \textbf{0.47} & \textbf{99.60} \\
\midrule
\multirow{2}{*}{Qwen3-VL-8B} & Baseline & 73.22 & 0.32 & 0.66 & 95.21 & 70.34 & 0.52 & 1.49 & 94.02 \\
& Ours & \textbf{95.40} & \textbf{0.05} & \textbf{0.21} & \textbf{100.00} & \textbf{93.40} & \textbf{0.07} & \textbf{0.26} & \textbf{100.00} \\
\midrule
\multirow{2}{*}{LLaVA-1.5-7B} & Baseline & 24.07 & 4.46 & 6.58 & 39.22 & 33.30 & 1.65 & 2.34 & 56.71 \\
& Ours & \textbf{75.30} & \textbf{0.26} & \textbf{0.53} & \textbf{98.90} & \textbf{72.50} & \textbf{0.29} & \textbf{0.56} & \textbf{98.70} \\
\bottomrule
\end{tabular}%
}

\caption{In-distribution counting benchmarks with full metrics.}
\label{tab:in_distribution_full}
\end{table*}
Across three backbones, our method consistently improves in-distribution counting on both SynDot and SynPoly. The gain is not only in exact accuracy, but also in the size of counting errors: MAE/RMSE drop clearly, which means mistakes become much smaller rather than only shifting between nearby counts. In addition, OBO increases to 100\% for the stronger models, showing that predictions are almost always within one count of the correct answer.

A notable pattern is that SynPoly benefits more than SynDot for the same backbone. SynPoly contains more varied shapes and layouts, so it is a harder in-distribution subset. The larger improvement there suggests the method helps with structured scenes and complex visual grouping, not only with simple dot patterns.

\subsection{Additional Counting Evaluation}
\label{sec:additional_counting}

\subsubsection{Counting Evaluation on CountBenchQA}

% requires: \usepackage{booktabs,multirow}
\begin{table}[t]
\centering
\small
\setlength{\tabcolsep}{8pt}
\renewcommand{\arraystretch}{1.15}
\caption{CountBenchQA results.}
\label{tab:countbenchqa}
\begin{tabular}{llcccc}
\toprule
\textbf{Model} & \textbf{Method} & \textbf{Acc} $\uparrow$ & \textbf{MAE} $\downarrow$ & \textbf{RMSE} $\downarrow$ & \textbf{OBO} $\uparrow$ \\
\midrule
\multirow{2}{*}{Qwen2.5-VL-7B} & Baseline & 80.44 & 0.66  & 4.34 & 89.61 \\
                              & Ours     & \textbf{82.08} & \textbf{0.58} & \textbf{4.28} & \textbf{91.24} \\
\midrule
\multirow{2}{*}{Qwen3-VL-8B}   & Baseline & 87.98 & 0.23   & 0.86 & 95.52 \\
                              & Ours     & \textbf{90.22} & \textbf{0.20} & \textbf{0.85} & \textbf{95.93} \\
\midrule
\multirow{2}{*}{LLaVA-1.5-7B}  & Baseline & 42.16 & 1.38   & 2.08 & 59.67 \\
                              & Ours     & \textbf{47.86} & \textbf{1.15} & \textbf{1.78} & \textbf{67.01} \\
\bottomrule
\end{tabular}
\end{table}

As shown in~\cref{tab:countbenchqa}, our method yields consistent improvements over the baseline across all three backbones on the real-world CountBenchQA benchmark~\cite{paiss2023teaching}. Notably, LLaVA-1.5-7B benefits the most, with accuracy increasing by $+5.70$ percentage points (from $42.16\%$ to $47.86\%$), MAE decreasing from $1.38$ to $1.15$, RMSE dropping from $2.08$ to $1.78$, and OBO rising substantially from $59.67\%$ to $67.01\%$. This large margin suggests that the weaker baseline has more room for improvement, and our interpretability-guided training effectively strengthens its visual counting circuits. For the stronger Qwen-series models, the gains are moderate but consistent: Qwen3-VL-8B achieves $+2.24$ accuracy improvement (from $87.98\%$ to $90.22\%$) while Qwen2.5-VL-7B improves by $+1.64$ points. Importantly, even for these already-strong baselines, all four metrics improve simultaneously, confirming that the enhancement is not limited to exact-match accuracy but extends to reducing error magnitude (MAE/RMSE) and improving near-miss reliability (OBO). These results demonstrate that the counting ability acquired from synthetic training transfers effectively to real-world images with natural backgrounds, complex textures, and diverse object categories.

\subsubsection{Counting Beyond 10}

To further verify that our method enhances intrinsic counting capabilities rather than inducing rote memorization, we expand the object range of the SynDot dataset to [1, 30]. Specifically, we train Qwen3 exclusively on the 1 to 10 range and evaluate its performance on two unseen intervals: 11 to 20 and 21 to 30. As shown in \cref{tab:range_performance}, our approach consistently improves overall counting accuracy across these extrapolated ranges. This confirms a genuine enhancement of the model's foundational numerosity mechanisms.

\begin{table}[ht]
\centering
\caption{Performance comparison over larger counting ranges for Qwen3-VL-8B. \textbf{Ours} are finetuned on the number range 1-10 only.}
\label{tab:range_performance}
\small
\begin{tabular*}{\linewidth}{@{\extracolsep{\fill}}lllcccc@{}}
\toprule
\textbf{Model} & \textbf{Method} & \textbf{Range} & \textbf{Acc} $\uparrow$ & \textbf{MAE} $\downarrow$ & \textbf{RMSE} $\downarrow$ & \textbf{OBO} $\uparrow$ \\ \midrule
\multirow{4}{*}{Qwen3-VL-8B} 
& \multirow{2}{*}{Baseline} 
& 11-20 & 35.00 & 0.90 & 1.25 & 80.02 \\
& & 21-30 & 10.42 & 2.49 & 3.02 & 32.37 \\
% \cmidrule{2-7}
& \multirow{2}{*}{Ours} 
& 11-20 & \textbf{48.16} & \textbf{0.59} & \textbf{0.86} & \textbf{93.45} \\
& & 21-30 & \textbf{18.27} & \textbf{1.61} & \textbf{2.05} & \textbf{52.81} \\
\bottomrule
\end{tabular*}
\end{table}

\subsection{Additional Ablation Study}
\label{sec:additional_ablation}

We provide additional ablation studies in this section for the layer choice of Attention Regularizer and the value choice of $\alpha$ for head temperature tuning.

\subsubsection{Layer Choice for Attention Regularizer}

\begin{figure*}[htbp] 
    \centering
    
    % (a) Qwen 2.5 Analysis
    \begin{subfigure}{\linewidth}
        \centering
        \includegraphics[width=\linewidth]{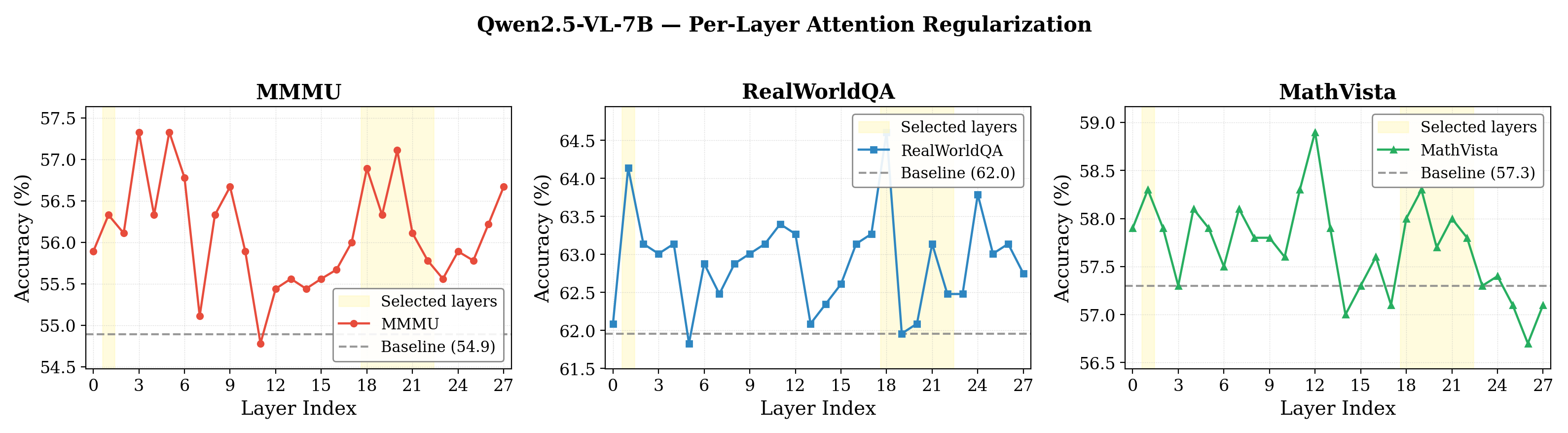}
        \caption{Layer-wise ablation analysis for \textbf{Qwen2.5-VL-7B}.}
        \label{fig:ablation_qwen25}
    \end{subfigure}
    
    % (b) Qwen 3 Analysis
    \begin{subfigure}{\linewidth}
        \centering
        \includegraphics[width=\linewidth]{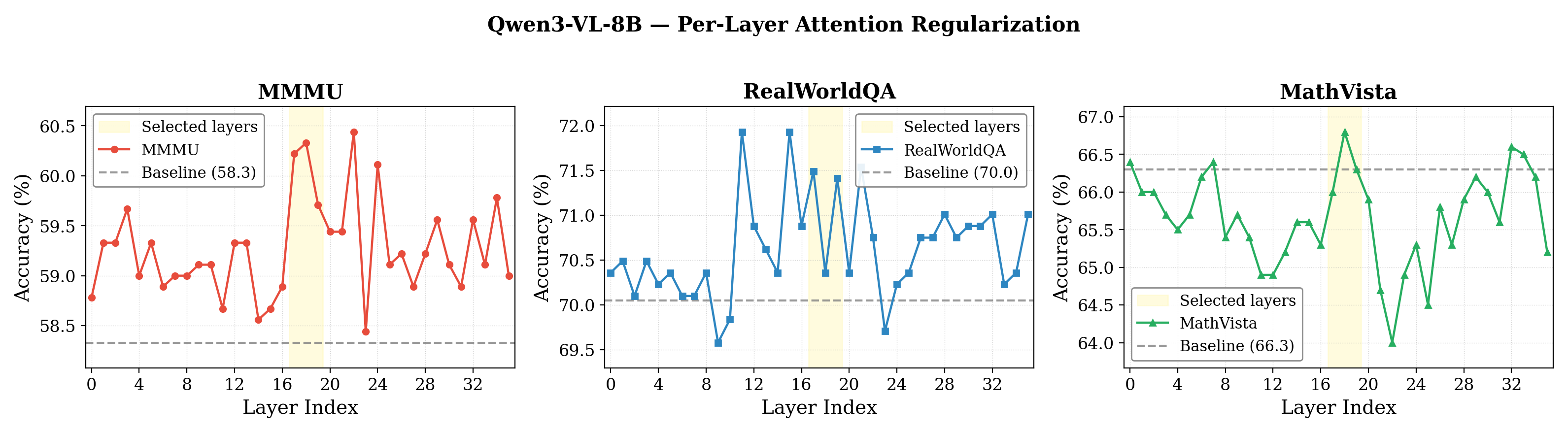}
        \caption{Layer-wise ablation analysis for \textbf{Qwen3-VL-8B}.}
        \label{fig:ablation_qwen3}
    \end{subfigure}

    % (c) LLaVA Analysis
    \begin{subfigure}{\linewidth}
        \centering
        \includegraphics[width=\linewidth]{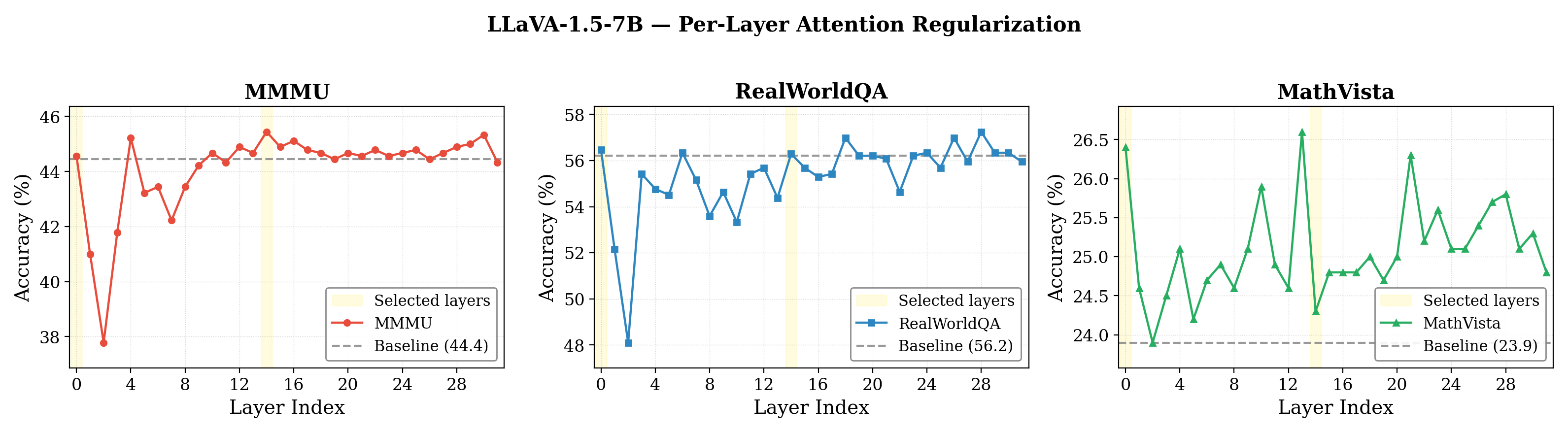}
        \caption{Layer-wise ablation analysis for \textbf{LLaVA-1.5-7B}.}
        \label{fig:ablation_llava}
    \end{subfigure}

    \caption{\textbf{Layer-wise ablation of the attention regularizer for three LVLMs.}}
    \label{fig:cross_model_ablation}
\end{figure*}

\cref{fig:cross_model_ablation} illustrates the performance variations across three visual reasoning tasks when applying the object focus regularizer to individual layers. We optimize the model using both SFT and $\mathcal{L}_{\text{focus}}$ on the SynDot and SynPoly datasets, isolating the regularization to a single layer per trial. Notably, applying this attention regularizer to the critical functional layers identified in our prior interpretability analysis consistently yields positive performance gains. This directly validates the practical utility of our mechanistic findings.

\subsubsection{Impact of $\alpha$ on Adaptive Head Temperature Tuning}

\begin{table}[t]
\centering
\small
\setlength{\tabcolsep}{6pt}
\renewcommand{\arraystretch}{1.08}
\caption{Ablation study of head temperature tuning (HTT) coefficient $\alpha$ and head importance reweighting (HIR) on Qwen2.5-VL-7B. The baseline method is SFT+$\mathcal{L}_{\text{focus}}$ with $\alpha=1.0$.}
\label{tab:ablation_htt}
\begin{tabular}{l c cccc}
\toprule
\textbf{Method} & \boldmath$\alpha$ & \textbf{MMMU} & \textbf{RealWorldQA} & \textbf{MathVista} & \textbf{Average} \\
\midrule
SFT + $\mathcal{L}_{\text{focus}}$ & 1.0 & 56.11 & 64.05 & 58.20 & 59.45 \\
+ HTT & 1.1 & {56.14} & 63.94 & 58.10 & {59.39} \\
+ HTT & 1.2 & 56.33 & 64.12 & 58.20 & 59.55 \\
+ HTT & 1.3 & \textbf{56.43} & 64.05 & 57.90 & 59.46 \\
\midrule
+ HTT + HIR & 1.2 & 56.33 & \textbf{64.14} & \textbf{58.30} & \textbf{59.59} \\
\bottomrule
\end{tabular}
\end{table}

We study the effect of the baseline inverse temperature coefficient $\alpha$ in Adaptive Head Temperature Tuning (HTT). Recall that each target head's logits are scaled by $\beta_h = \alpha \times \gamma_h$, where $\gamma_h$ is the head importance score from our circuit analysis. When $\alpha=1.0$, the attention distribution remains unmodified, reducing to the SFT+$\mathcal{L}_{\text{focus}}$ baseline. We sweep $\alpha \in \{1.1, 1.2, 1.3\}$ and additionally evaluate whether combining HTT with Head Importance Reweighting (HIR)---which re-scales the output contributions of the identified counting heads proportionally to their importance scores---yields further gains. All experiments are conducted on Qwen2.5-VL-7B and evaluated on three general visual reasoning benchmarks (MMMU, RealWorldQA, MathVista) to assess whether sharpening target heads preserves or improves general capabilities beyond counting.

As shown in~\cref{tab:ablation_htt}, moderate temperature sharpening ($\alpha=1.2$) yields the best trade-off, improving the average score from $59.45$ to $59.55$ with consistent gains on MMMU ($+0.22$) and RealWorldQA ($+0.07$) without any degradation on MathVista.
A milder scaling ($\alpha=1.1$) produces negligible change, suggesting that the entropy reduction is too small to meaningfully amplify the counting circuit.
A more aggressive scaling ($\alpha=1.3$) boosts MMMU to the highest single-benchmark score ($56.43$) but causes a slight decline on MathVista ($-0.30$), indicating that over-sharpening attention may suppress non-counting heads that contribute to mathematical reasoning. This highlights a precision--generality trade-off: excessively concentrating attention on counting-critical heads risks narrowing the information bandwidth available for other reasoning pathways.

Combining HTT ($\alpha=1.2$) with HIR achieves the best overall performance ($59.59$ average), simultaneously improving all three benchmarks. This validates that the two interventions are complementary: HTT sharpens \emph{where} the counting heads attend, while HIR amplifies \emph{how much} their outputs contribute to the residual stream, together strengthening the signal-to-noise ratio of the counting circuit without harming general capabilities.

% \section{HeadLens VS AttentionLens}

\section{Control Experiment: Visual Recognition Training Does Not Improve Visual Reasoning}
\label{sec:color_shape_control}

A central finding of our work is that enhancing counting ability generalizes to broader visual reasoning benchmarks (\cref{tab:general_eval_clean}). A natural question arises: \textit{is this transfer a generic property of any visual fine-tuning, or is it specific to counting as a reasoning-intensive task?} To answer this, we conduct a controlled experiment using a pure visual recognition task---color and shape identification---that requires perception but minimal reasoning.

\subsection{Setup}

\noindent\textbf{Data.}
We construct a synthetic color-shape recognition dataset (\textbf{SynColorShape}) following the same generation protocol as our counting datasets. Each image is a $336\times336$ white canvas containing a single filled polygon drawn at a random position, random radius (30--80\,px), and random rotation. We use 10 colors (red, blue, green, yellow, orange, purple, pink, cyan, brown, gray) $\times$ 8 shapes (triangle, square, pentagon, hexagon, octagon, circle, star, diamond) = 80 combinations, uniformly sampled to produce 4{,}960 training samples (62 per combination). Each sample is paired with either a color or shape question (50/50 split), and all ground-truth answers are single tokens.

\begin{figure}
    \centering
    \includegraphics[width=0.75\linewidth]{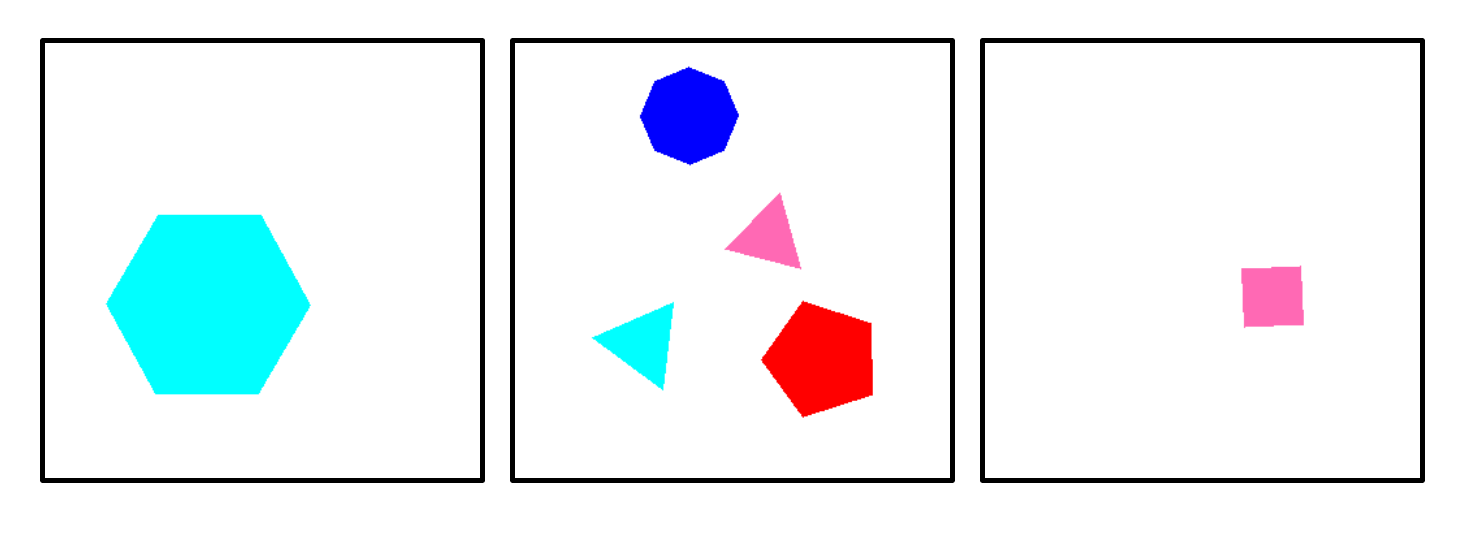}
    \caption{Data sample of SynColorShape.}
    \label{fig:color_shape}
\end{figure}

\noindent\textbf{Task and Prompts.}
The color-shape task asks the model to identify a single visual attribute of one object. Three prompt templates are randomly assigned per sample for each task type:

\noindent\textit{Color recognition:}
\begin{itemize}
    \item ``What is the color of the \{shape\} in the image? Answer with the color name only.''
    \item ``Identify the color of the \{shape\} shown. Answer with the color name only.''
    \item ``What color is the \{shape\} in this image? Answer with the color name only.''
\end{itemize}

\noindent\textit{Shape recognition:}
\begin{itemize}
    \item ``What is the shape of the \{color\} object in the image? Answer with the shape name only.''
    \item ``Identify the shape of the \{color\} figure shown. Answer with the shape name only.''
    \item ``What shape is the \{color\} object in this image? Answer with the shape name only.''
\end{itemize}

\noindent All prompts end with a constraint phrase to enforce single-word answers. We apply the same answer-only supervision strategy as our counting SFT: cross-entropy loss is computed exclusively on the 1--2 answer tokens, with all image, prompt, and special tokens masked.

\noindent\textbf{Training.}
We fine-tune Qwen2.5-VL-7B-Instruct using LoRA ($r\!=\!16$, $\alpha\!=\!32$) on all linear layers with the same optimizer setting (AdamW, $\mathrm{lr}\!=\!2\!\times\!10^{-5}$, cosine schedule). We additionally train a second-stage attention regularization model (LoRA $r\!=\!8$, $\alpha\!=\!16$) that applies KL-divergence supervision on the top-20 important heads identified via activation patching, pushing their attention toward the object mask. This mirrors the two-stage pipeline used in our counting method.

\subsection{Results}

The color-shape SFT dramatically improves the target recognition task: shape accuracy rises from 78.1\% to 99.8\% (with diamond going from 0\% to 100\%), and color accuracy reaches a perfect 100\%. However, the critical observation is its effect on general reasoning benchmarks:

\begin{table}[h]
\centering
\small
\setlength{\tabcolsep}{5pt}
\renewcommand{\arraystretch}{1.08}
\caption{Change in general reasoning benchmarks ($\Delta$ from each method's own baseline) for Qwen2.5-VL-7B. Color-shape recognition training shows no positive transfer, while counting training yields consistent gains.}
\label{tab:color_shape_control}
\begin{tabular}{lcccc}
\toprule
Training Task & $\Delta$MMMU & $\Delta$MathVista & $\Delta$RealWorldQA & $\Delta$ (avg.) \\
\midrule
Color-Shape SFT & $-$0.1 & $-$0.7 & $-$0.2 & $-$0.3 \\
Color-Shape SFT + AttnReg & $-$0.2 & $-$0.8 & +0.2 & $-$0.3 \\
Counting (Ours) & \textbf{+1.44} & \textbf{+1.00} & \textbf{+2.18} & \textbf{+1.54} \\
\bottomrule
\end{tabular}
\end{table}

Despite successfully learning the target visual recognition task, color-shape training produces \textit{no positive transfer} to general reasoning benchmarks. MMMU, MathVista, and RealWorldQA all remain within measurement noise or show slight degradation ($\Delta_\text{avg} \approx -0.3\%$). By contrast, our counting-based training with the same LoRA fine-tuning paradigm achieves a consistent $+1.54\%$ average improvement across the same benchmarks.

\subsection{Analysis}

This control experiment provides direct evidence that the generalization observed with counting training is \textbf{not} a generic byproduct of visual fine-tuning. While both tasks involve synthetic images, single-token answers, and the same training pipeline, only counting training transfers to broader reasoning. This aligns with our mechanistic analysis in \cref{fig:cross_task}: the Jaccard similarity between \textit{Attribution} (color/shape) and reasoning tasks is remarkably low, confirming that recognition relies on perception-only circuits with minimal overlap to the shared reasoning sub-networks.

The key distinction is that counting requires \textit{visual reasoning}: the model must individuate objects, maintain a running tally, and abstract a numerical quantity, operations that engage cross-modal routing and aggregation heads shared with other reasoning tasks. In contrast, color and shape recognition are fundamentally perceptual; they require identifying a visual attribute of a single object, which is largely resolved by early-layer feature extraction without engaging the deeper reasoning circuitry.

This finding strengthens our central claim: counting serves as a uniquely effective proxy task for enhancing visual reasoning in LVLMs precisely because it activates and refines the shared computational circuits underlying general visual understanding.

\section{Case Study}

We visualize the attention maps of the baseline and our proposed method to gain insights into the model's counting behavior.
For each case, we plot (i) the visual attention distribution at Layer 2 (the layer where we apply \textit{Object-Focused Attention Regularizer} and \textit{Adaptive Head Temperature Tuning}) and (ii) the attention distribution averaged over all layers.

\begin{figure}[h]
  \centering
  \includegraphics[width=1.0\linewidth]{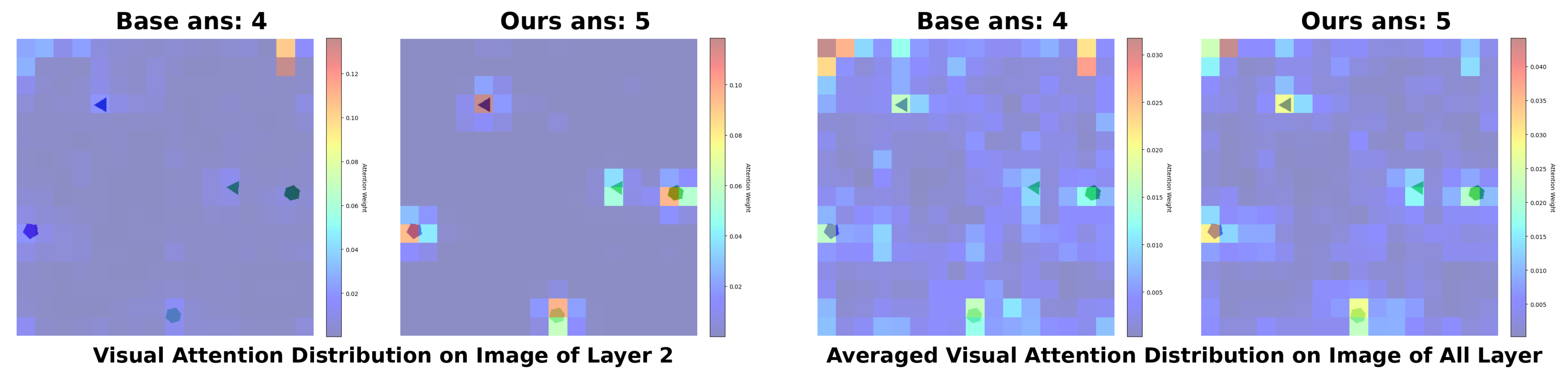}
  \caption{SynPoly case studies at GT=5. \textbf{Prompt:} How many colorful polygons are in the image? Answer with only an integer. \textbf{Observation:} Base predicts 4 and Ours predicts 5. Base attention is more spread with weak peaks, while ours shows sharper peaks aligned with the dots already at Layer 2 and remains aligned in the all-layer average. \textbf{Conclusion:} Our method strengthens early visual grounding for counting, making the attention more instance-aligned even when both models output the correct number.}
  \label{fig:case_polygon_gt5}
\end{figure}

\begin{figure}[h]
  \centering
  \includegraphics[width=1.0\linewidth]{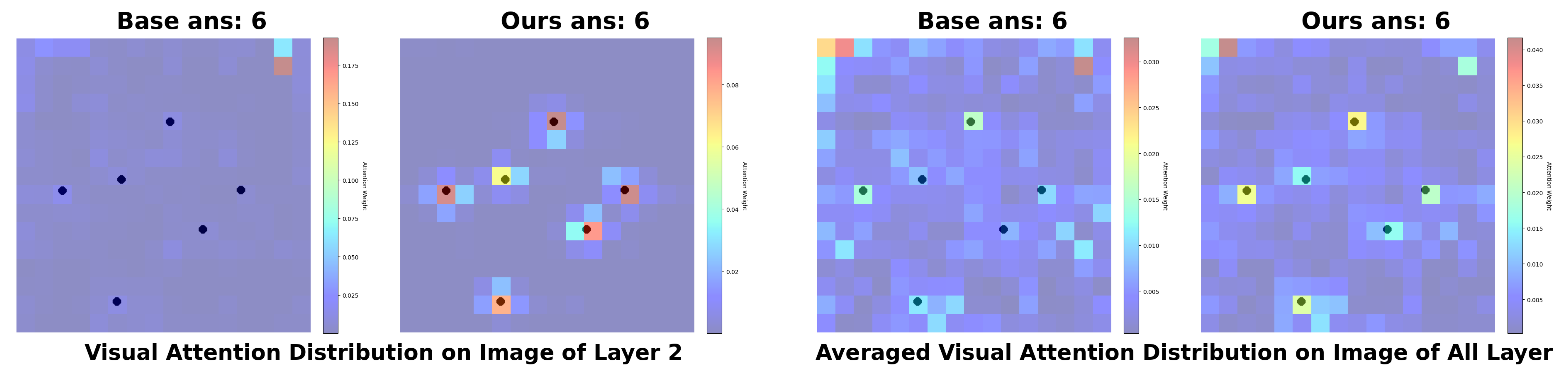}
  \caption{SynDot case studies at GT=6. \textbf{Prompt:} How many black dots are in the image? Answer with only an integer. \textbf{Observation:} Base predicts 6 and Ours also predicts 6. However, base attention is more spread with weak peaks, while ours shows sharper peaks aligned with the dots already at Layer 2 and remains aligned in the all-layer average. \textbf{Conclusion:} Our method strengthens early visual grounding for counting, making the attention more instance-aligned even when both models output the correct number.}
  \label{fig:case_black_dots_gt5}
\end{figure}

%%%%%%%%%%%%%%%%%%%%%%%%%%%%%%%%%%%%%%%%%%%%%%%%%%%%%%%%%%%%

\end{document}